\documentclass[default]{sn-jnl}

\usepackage{xurl}   % allows URL line breaks at any character
\usepackage{graphicx}%
\usepackage{multirow}%
\usepackage{amsmath,amssymb,amsfonts}%
\usepackage{amsthm}%
\usepackage{mathrsfs}%
\usepackage[title]{appendix}%
\usepackage{xcolor}%
\usepackage{textcomp}%
\usepackage{fancybox}
\usepackage{placeins}
\usepackage{manyfoot}%
\usepackage{booktabs}%
\usepackage{algorithm}%
\usepackage{algorithmicx}%
\usepackage{algpseudocode}%
\usepackage{listings}%
\usepackage{adjustbox}%
\usepackage{cleveref}%
\usepackage{tabularx}%
\usepackage{tikz}
\usepackage{subcaption} % For creating subfigures
\usepackage{caption}
\usepackage[shortlabels]{enumitem}
\usepackage{fancyhdr}
\DeclareCaptionLabelSeparator{bar}{ | }

\newcounter{edfig}
\newcounter{tmpfig}

\newenvironment{edfigure}[1][tbp]{%
  \setcounter{tmpfig}{\value{figure}}%            stash the main counter
  \begin{figure}[#1]%
    \setcounter{figure}{\value{edfig}}%           continue ED numbering
    \captionsetup{name=Extended Data Fig., labelsep=bar}%
}{%
    \setcounter{edfig}{\value{figure}}%           remember where ED got to
  \end{figure}%
  \setcounter{figure}{\value{tmpfig}}%            restore the main counter
}

\newcommand{\edref}[1]{Extended Data Fig.~\ref{#1}}

\DeclareCaptionFont{custom}{\fontsize{13.5}{14}\selectfont}
\usepackage{array}
\usepackage{arydshln}
\definecolor{olive}{rgb}{0,0.6,0}

\theoremstyle{plain}

\theoremstyle{remark}
\newtheorem{rem}{\protect\remarkname}
\theoremstyle{definition}

\usepackage{babel}
\providecommand{\definitionname}{Definition}
\providecommand{\remarkname}{Remark}
\providecommand{\theoremname}{Theorem}

\usepackage{mathtools}
\DeclareMathAlphabet\mathbfcal{OMS}{cmsy}{b}{n}
\usepackage{bm}
\usepackage{amsthm}

\theoremstyle{thmstyleone}%
\theoremstyle{thmstyletwo}%

\theoremstyle{thmstylethree}%
\usepackage{bm}

\makeatletter
\renewcommand{\@maketitle}{\newpage\null%
    \if@remarkboxon\vbox to 0pt{\vspace*{-78pt}\hspace*{-18pt}\FMremark}\else\vskip10pt\fi%%\par%
    \hsize\textwidth\parindent0pt%%%\vskip7pt%
    {\hbox to \textwidth{{\Artcatfont\ArtType\hfill}\par}}
    \ifx\@title\empty\else%
        \removelastskip\vskip10pt\nointerlineskip%
        {\Titlefont\@title\par}
    \fi%
    \ifx\@subtitle\empty\else%
        \vskip9pt%
        {{\SubTitlefont\@subtitle\par}}
    \fi%
    \ifnum\aucount>0
        \global\punctcount\aucount%
        \vskip10pt%
        \artauthors\par%%     authors and emails
        {\vskip4pt\addressfont\auaddress\par%%      corresponding address
	 \removelastskip\vskip8pt%
	\ifnum\emailcnt>0\relax%
           \ifx\corrauthemail\@empty\else{\ifnum\aucount>1*\fi}%
	   Corresponding author(s). E-mail(s): \corrauthemail\par\fi%
	   \ifx\authemail\@empty\else Contributing authors:\ \authemail\fi%
        \fi%
        \ifequalcont{\par$^{\dagger}$\@equalconttext\par}\fi%
	 \removelastskip\vskip8pt%
        \ifpresentaddress{\par\@presentaddresstext\par}\fi%
	}
     \fi%
     {\printabstract\par}%
     {\printkeywords\par}%
     \ifx\@pacs\empty\else%
       \loop\ifnum\PacsCount>0%
          \csname\romannumeral\PacsTmpCnt StorePacsTxt\endcsname\par%
          \StepDownCounter{\PacsCount}%
          \StepUpCounter{\PacsTmpCnt}%
       \repeat%
    \fi%
    \removelastskip\vskip14pt\vskip0pt}%
\makeatother
\begin{document}

\title{Accelerated Reinforcement Learning for Soft Robotics: Hours Instead of Years}


\author[1]{\fnm{Roshan S.} \sur{Kaundinya}}\equalcont{These authors contributed equally to this work.}

\author[2]{\fnm{Hehui} \sur{Zheng}}\equalcont{These authors contributed equally to this work.}

\author*[2]{\fnm{Robert K.} \sur{Katzschmann}}\email{rkk@ethz.ch}

\author*[1]{\fnm{George} \sur{Haller}}\email{georgehaller@ethz.ch}

\affil[1]{\orgdiv{Institute for Mechanical Systems}, \orgname{ETH Z\"urich}, \orgaddress{\street{Leonhardstrasse 21}, \postcode{8092}, \city{Z\"urich}, \country{Switzerland}}}

\affil[2]{\orgdiv{Soft Robotics Lab}, \orgname{ETH Z\"urich}, \orgaddress{\street{Tannenstrasse 3}, \postcode{8092}, \city{Z\"urich}, \country{Switzerland}}}

%%==================================%%
%% unstructured abstract            %%
%%==================================%%

\abstract{
Reinforcement learning (RL) requires large amounts of interaction data or a reliable simulator, both hard to obtain for nonlinear soft robotic hardware. Recent work shows that a time-scale separation inherent to soft robot dynamics justifies their reduction to low-dimensional invariant manifolds, known as adiabatic spectral submanifolds (aSSMs). We demonstrate that an aSSM-reduced model, a simple polynomial ordinary differential equation learned purely from pre-collected data, can serve as an RL training environment. On a pneumatically actuated soft arm, we train a trajectory-tracking policy in 3 hours on a six-dimensional aSSM-reduced model and deploy it without fine-tuning, whereas reproducing this policy training extrapolates to 3 years of hardware interaction. Our policy tracks dynamic targets with 40\% lower error than otherwise identical policies trained on a 4.8-million-parameter recurrent neural network (RNN) or a 1.7-million-parameter transformer, which take 13 and 100 hours to identify, against 5 minutes for our mathematically justified 6,924-parameter model.}

%Reinforcement learning (RL) requires either large amounts of interaction data or a reliable simulator to learn effective control policies, both of which are hard to obtain for nonlinear soft robotic hardware. Recent work has shown that a time-scale separation inherent to soft robot dynamics justifies their reduction to low-dimensional invariant manifolds, known as adiabatic spectral submanifolds (aSSMs). We demonstrate that an aSSM-reduced model, a simple polynomial ordinary differential equation learned purely from pre-collected data, can be used as the training environment to learn RL policies for soft robotic hardware. 
%Specifically, on a pneumatically actuated soft arm, we learn an RL policy on a six-dimensional aSSM-reduced model for trajectory tracking, and deploy it on hardware without fine-tuning. Reproducing this RL policy training directly on the robot requires 3 years of continuous interaction, whereas training on the reduced model takes 3 hours. Our aSSM-based policy outperforms otherwise identical policies trained on recurrent neural network (RNN) and transformer-based dynamics models on dynamic target-tracking tasks. These results indicate that for soft robots, a mathematically justified aSSM-reduced model serves as a significantly faster and more accurate training environment for RL than a large, black-box neural network, and a far more practical one than the delicate hardware itself.
% reduction in number of parameters and also training time saved are left out in the abstract. 


\maketitle

\clearpage

%Point 0.5- General intro about RL in robotics.

Learning control policies for robots to excel at human- or animal-oriented skills, such as walking, swimming, flying, and playing physical or cognitive games, is an ambitious open problem of current interest in the machine learning and robotics communities (see \citet{peters_review} and \citet{levine_review} for a review). Since biological morphologies and their associated mechanics heavily inspire robot designs, the development of general-purpose controllers that emulate the trial-and-error learning processes of biological organisms to master complex skills has received broad interest.

% What is RL
Reinforcement learning (RL) formalizes control by trial and error (see \citet{sutton2018reinforcement}). 
An RL agent learns the control policy that maps an observation of the robot and the current target to a distribution over control actions. 
Control actions are sampled from the policy and applied to the robot. 
The resulting state is observed, and the policy parameters are updated to maximize a reward that quantifies the proximity to the target associated with the skill.
% Learning a generalizable RL policy requires extensive exploration of the state--action space of the robot and its environment. 

% SOTA model-free + simulator
Methods that learn from rewards evaluated on interaction data, the model-free family, require significantly more data than a physical robot can usually provide, since every environment step costs wall-clock time and mechanical life of the hardware.
Developments in rigid robotics alleviated this by moving the exploration off the hardware into high-fidelity, parallelizable simulators (see \citet{makoviychuk2021isaac}), an approach that produced the locomotion (see \citet{locomotion_review}) and manipulation (see \citet{manipulation_review}) results now state of the art on a variety of rigid robotic platforms such as ANYmal and the Franka arm.

% Simulators for soft robots bad
The same pipeline is generally not available for soft robots.
For certain types of soft robotic hands (see \citet{yasu_2023}) and continuum arms (see \citet{satheeshbabu2020}), rigid-body physics engines such as MuJoCo, or elastic rod-based simulators (see \citet{naughton21}) that approximate their kinematics already aid in training model-free RL policies.
But they suffer from the simulation-to-reality gap for dexterous manipulation tasks.
On the other hand, current soft robotic simulators that are able to capture the nonlinear dynamics (see \citet{faure2012sofa}, \citet{sora_26}, and \citet{elastica_pp_2026}) are computationally costly to be integrated in the policy training loop. 
These practical limitations motivate the development of learning RL policies directly on hardware soft robot data.

Attempts to control soft robots through an RL policy trained solely on hardware data are not new. The earliest work by \citet{nath2007} learned a model-free policy to control a soft robotic manipulator. The method, however, does not appear scalable to complex soft-robot geometries. Indeed, even with advances in RL algorithms and computational power, no successful model-free policies for complex soft robots have emerged to date. The reasons are not surprising, as model-free policies require extensive hardware exploration to converge, which, for delicate soft robots, is nearly impossible because their softness, coupled with repeated use, leads to memory effects and plasticity that tend to cause hardware failure.

%(often termed model-free as they large models they learn are assumed to resemble simulators)
At the crossroads of RL implementation difficulties: dealing with delicate hardware and no access to a reliable, fast simulator. Large black-box neural network-based models trained on dynamic hardware data to model the dynamics of soft robots offer a solution. The underlying hope is that the neural network's large parameter count provides a reasonable estimate of the soft robot's high fidelity, and the more appealing aspect is that neural-network-based control policies are largely compatible with training on such models, ensuring fast, parallelizable training that explores the model's state space. These are termed model-based RL methods and are further sub-classified as online when policy training happens directly by interacting with hardware, or offline when policy training happens using the model as a training environment.

On soft robotic manipulators, \citet{Thuruthel2019ModelBasedRL} learns a nonlinear autoregressive model of the dynamics using a one-step memory window. They then employ this model inside a classical model predictive controller to generate target trajectories, on which an RL agent learns a control policy while interacting with the hardware. Recent work by \citet{zheng2025learningsoftroboticdynamics} and \citet{fang_2026} uses a larger neural network with one-step memory information and trains simultaneously with a control policy on hardware data. These online model-based methods have seen success, but, as their authors point out, it is unclear whether the errors can be attributed to the policy or to model learning (see \citet{mbrl_trust}). All in all, these approaches, though data-efficient, still converge to RL policies that are task-specific and need additional interaction with the hardware to be generally applicable.

To mitigate these issues, \citet{rl_aided} and \citet{koopman_RL} focus on training a long short-term memory (LSTM) and a multilayer perceptron (MLP), respectively, both with larger memory windows, and then train control policies on these models. However, they report accuracy gaps when applied to the hardware and hence use online model-free RL methods to fine-tune their predictions. For purely offline model-based methods for soft robots, \citet{uljad_mbrl} and \citet{mbrl_soft_robotbest} use LSTMs to learn the dynamics of soft arms and train policies on them. Notably, the latter tracks dynamic targets using an LSTM-based RL policy, achieving accuracy on planar trajectories. Their precollected data for training the LSTM is modest. Still, they offer no systematic guidelines for choosing the network architecture, hyperparameters, and memory windows, which are essential for LSTMs, which are prone to long-term prediction inaccuracies and poor generalization to unseen test data (see \citet{kaundinyaGSS2026}). 
\begin{figure}[H]
        \includegraphics[width=\linewidth]{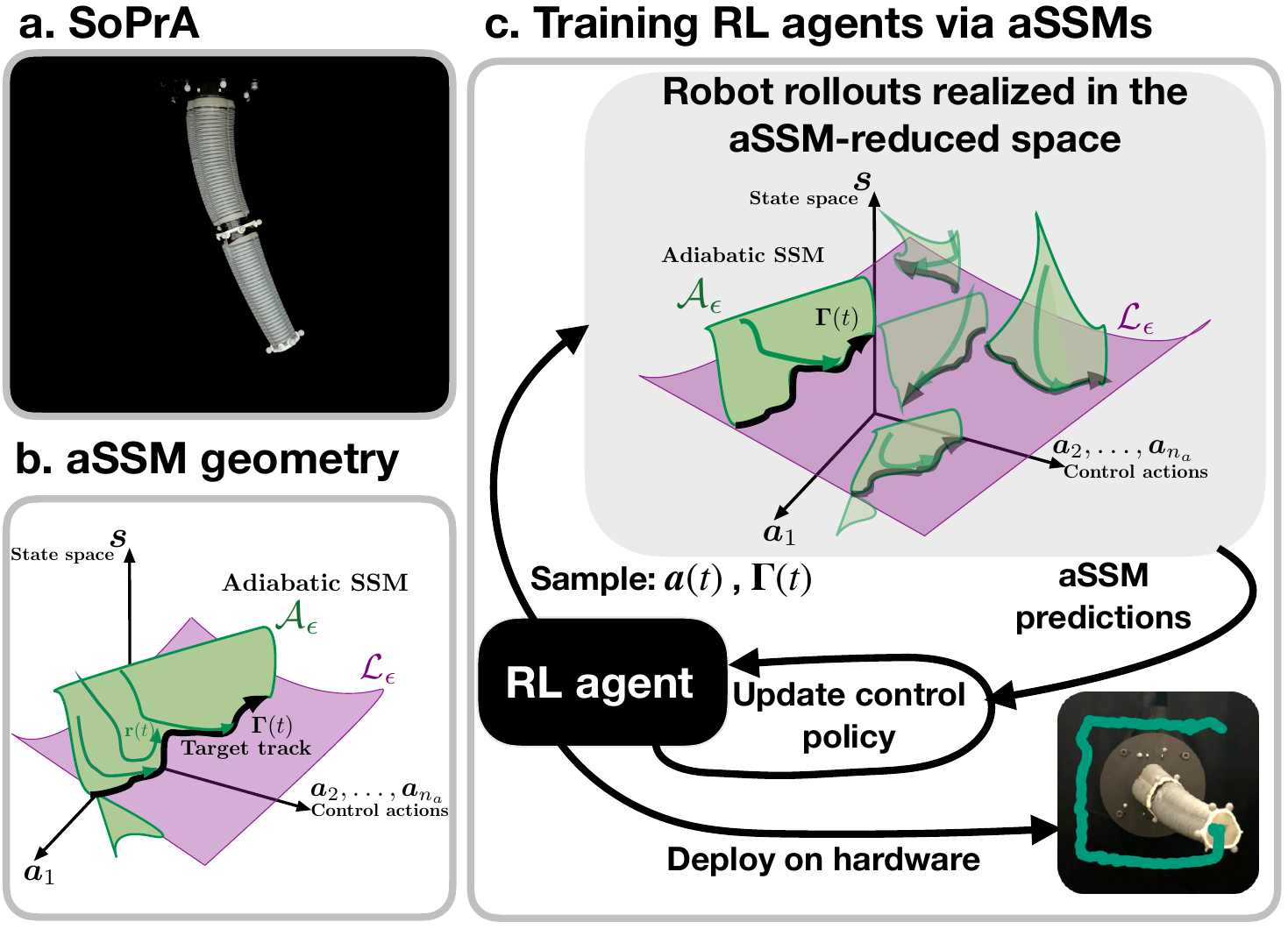}  
    \caption{\textbf{[Overview of learning an aSSM RL policy on a soft robot]}. (a) Snapshot of the SoPrA arm. (b) Geometry of the adiabatic spectral submanifold (aSSM) in the state- and control action- space. (c) General overview of training an RL agent on an aSSM-reduced model.}
    \label{fig:scheme}
\end{figure}

As an alternative to data-driven black-box modeling, the mathematical theory of spectral submanifolds (SSMs), which are low-dimensional attracting invariant manifolds (see \citet{haller25} for a review), provides a rigorous geometric understanding of the dominant dynamics of nonlinear dynamical systems. Specifically, restricting the dynamics to slow SSMs allows for justified model reduction of nonlinear continuum mechanical systems described by ordinary differential equations (ODEs) (see \citet{haller16} and \citet{haller23}) or partial differential equations (PDEs) (see \citet{buza2026spectralsubmanifoldreductionpdes}). The resulting SSM-reduced model is a low-dimensional ODE, which can be extracted purely from data as a simple polynomial model (see \citet{cenedese21} and \citet{axas22}), and has demonstrated predictive accuracy in a range of mechanical experiments subject to previously unseen external forcing (see \citet{cenedese22a}, \citet{cenedese22b}, \citet{Bettini2025}, \citet{Bettini2026}, and \citet{morsy2024reducingfiniteelementmodels}).

Dominant dynamics of nonlinear soft robots captured by slow SSMs have already been locally extracted near equilibria and used for controls in \citet{alora24} for a hardware Diamond robot. Only recently, in \citet{Kaundinya2026}, a time-scale separation phenomenon was identified as common to all soft robotics control operations, enabling the global extraction of the dominant dynamics onto adiabatic spectral submanifolds (aSSMs) via extensions to the local theory of SSMs (see \citet{haller24_wa}). They show, across a variety of soft robotic simulators, that aSSMs extracted from data can be used in a model predictive control (MPC) scheme successfully for accurate trajectory tracking. 

This methodology has very recently been demonstrated on hardware, by further reducing aSSMs onto slow manifolds to control artificial muscles (see \citet{Bettini2026_muscles}) and to precisely control helicoid soft robots along 3D target tracks (see \citet{kaundinya2026precision}). Existing aSSM-based controllers pair the reduced model with classical numerical control techniques, which linearize the dynamics at each solve step. As a consequence, the nonlinear structure of the aSSM model is only partially exploited. Control methods, such as an RL agent optimizing policies directly on nonlinear model rollouts, without repeated linearization, have therefore remained unexplored for aSSM-reduced models. We will fill this gap with the present work.

\textbf{Contribution:} We develop a method to train RL control policies offline on a data-driven aSSM-reduced model (see Fig.~\ref{fig:scheme}c), without the RL training interacting with the hardware robot. Specifically: 
\begin{itemize}
    \item \textbf{[aSSM for a new hardware platform]} We extract and learn a six-dimensional aSSM-reduced model for a pneumatically actuated soft robot, a soft continuum proprioceptive arm (SoPrA)  (see Fig.~\ref{fig:scheme}a and Section~\ref{sec:softarm}), purely from hardware data. Previously, the data-driven aSSM methodology has been tested on tendon-driven trunk robots (see \citet{kaundinya2026precision}). This demonstrates the generality of our method, which is agnostic to the physics of the actuating mechanics.   
    \item \textbf{[Pareto dominant aSSM-RL policies]} Combining traditional RL policy optimization (see \citet{schulman2017proximalpolicyoptimizationalgorithms}) with our aSSM-reduced model (see Section~\ref{sec:aSSM-RL}) leads to Pareto dominant aSSM-RL control policies across model complexity, model learning time, RL policy training time, and closed-loop accuracy (see Section~\ref{sec:pareto-doms}). We use baselines that are state-of-the-art autoregressive neural network-based models from \citet{li2025robotic}, which they term robotic world models (RWMs). 
    \item \textbf{[Faster control]} Our aSSM-RL policy operates at a controller frequency of $100$ [Hz], which provides a $ 100$x speed-up over neural-network-based MPC controllers in \citet{zheng2025learningsoftroboticdynamics} and a $ 2$x speed-up over aSSM-based MPC controllers in \citet{kaundinya2026precision}. 
    \item \textbf{[Improved closed-loop accuracy compared to prior work]} Our aSSM-RL policy reduces closed-loop tracking error by $40\%$ relative to the best-performing autoregressive model-based baseline, across all trajectory tracking tasks (Section~\ref{sec:closedloop}), and by $20\%$ relative to our own aSSM-MPC baseline. Furthermore, aSSM-RL reduces tracking error by $55\%$ relative to the LSTM-based policy of \citet{mbrl_soft_robotbest} and by $76\%$ relative to the online model-based policies of \citet{fang_2026}.
\end{itemize}

\section{Results}\label{sec:results}
We present a detailed case study on the performance of model-based RL policies for controlling a soft continuum arm. Specifically, our experimental platform is the SoPrA arm (see \citet{sopra2021} for details), a pneumatically actuated soft continuum arm. The SoPrA arm is mounted vertically and hangs against gravity (see Fig.~\ref{fig:scheme}a). The cross-sectional radius of the arm at the free end is $2$ [cm] and $3$ [cm] at the base. It has a longitudinal length of $30$ [cm]. The arm is assembled by putting together two modules. Each module is cast in silicone material and comprises three expandable chambers. The chambers are connected to a pressure-driven actuation system that sets the pressure in each chamber to control the extent of expansion. We refer to the Supplementary Information for further hardware details.  

An external motion capture system monitors the arm's movements. The camera system records the positions of the free end, which we refer to as the tip, and the mid-section, which we refer to as the mid, where the modules meet. Our objective is to learn a model of the soft arm's dynamics using data collected by the camera system and the control actions applied to the robot to generate it. We then use this learned model to explore the soft arm's control action space and learn an RL control policy for target tracking tasks. 

On the modeling side, in Section \ref{sec:softarm}, we provide a detailed discussion of learning a low-dimensional 6D aSSM-reduced model for the arm. We also learn various, routinely used neural-network-based dynamics models for the arm's end-effector: multi-layer perceptrons (MLPs), recurrent neural networks (RNNs), and Transformers. Specifically, the RNN and Transformer baselines are the current state-of-the-art in recent work by \citet{li2025robotic}, which they denote as robotic world models (RWMs). We systematically compare the predictive power of all these models, including our aSSM-reduced model, for previously unseen random control actions in Section \ref{sec:black-box_model_comparisons}. The exact training details for all the neural network models used for comparisons are provided in the Supplementary Information.  

We then use our learned aSSM model in Section \ref{sec:aSSM-RL} to train a model-based RL policy for closed-loop control of the soft arm and follow this up in Section \ref{sec:pareto-doms} by outlining the Pareto dominance of using an aSSM-based RL policy compared to other neural-network-based policies. We finally deploy our aSSM-based RL policy on the hardware arm in Section \ref{sec:closedloop} to perform closed-loop tracking of 2D targets. 

\subsection{Data-driven modeling of the soft robotic arm via aSSMs}\label{sec:softarm}

As a continuum, the SoPrA soft arm is an infinite-dimensional dynamical system. Under soft-material constitutive laws and elastic continuum assumptions, the system's continuum equations of motion can be discretized to construct high-fidelity, high-dimensional dynamical systems. However, there are no robust high-fidelity simulators that faithfully model the hardware. Thus, we must rely only on data from a partially observable state space to infer a dynamical model for the SoPrA arm.

Under normal operating condition, the SoPrA arm exhibits an underlying time-scale separation: its intrinsic vibration rates die out significantly faster than the rate of change of control actions generated by the actuation system. Evidence of this time-scale separation has recently been documented in soft robotic hardware: the artificial muscle joint system of \citet{Bettini2026_muscles} and the soft helicoid trunk robot of \citet{kaundinya2026precision}. 

Relevant to our setting, the theoretical work by \citet{haller24_wa} extended to the control action setting by \citet{Kaundinya2026} provides proofs for the existence of a family of low-dimensional invariant manifolds called adiabatic spectral submanifolds (aSSMs), in the phase space of dynamical systems satisfying the time-scale separation assumption outlined above. Specifically, for target-tracking control, normally attracting aSSMs, $\mathcal{A}_{\epsilon}$, arise as the nonlinear continuations of the collection of dominant slow eigenspaces of the linearization of the robotic system about the frozen-time (i.e., $\epsilon =0$) limit of a target trajectory $\boldsymbol{\Gamma}(\epsilon t)$ (see Fig.~\ref{fig:scheme}b). An arbitrary initial condition in the phase space of the robot will converge quickly onto the aSSM and then later converge to the target trajectory. Hence, restricting the dynamics to $\mathcal{A}_\epsilon$ provides an accurate reduced model of the soft robot that can be used for controls. 

\citet{Kaundinya2026} and \citet{kaundinya2026precision} provide rigorous, data-driven methods for extracting low-dimensional, normally attracting aSSMs from partially observable state space coordinates of soft robots. We will closely follow their setup but use notation consistent with standard conventions in the Reinforcement Learning (RL) literature (see \citet{sutton2018reinforcement}). We define the partially observable state space of the arm as 
\begin{equation}
    \boldsymbol{s} \in \mathcal{S} \subseteq \mathbb{R}^{n_s},
\end{equation}
and the finite-dimensional control action,
\begin{equation}
    \boldsymbol{a} \in \mathcal{C} \subseteq \mathbb{R}^{n_a},
\end{equation}
where $\mathcal{S}$ and $\mathcal{C}$ denote the physically admissible state and control action subsets, respectively. In our current hardware setup for the SoPrA arm, we can only observe the tip and mid positions. Hence we have $n_s = 6$, and the observable state vector takes the form
\begin{equation}
\boldsymbol{s} = \{x,y,z\} = \{x_{\mathrm{tip}},x_{\mathrm{mid}}, y_{\mathrm{tip}},y_{\mathrm{mid}},z_{\mathrm{tip}},z_{\mathrm{mid}}\} \in \mathbb{R}^6.
\end{equation}
The pressure-driven actuation system of SoPrA has 6 air chambers, but due to current hardware limitations, only 5 are active and available to the user, which sets $n_a = 5$. 

\begin{figure}[t]
    \centering
    % Column 1: (a) step responses, (c) periodic state-space response, stacked
    \begin{subfigure}[t]{0.46\textwidth}
        \centering
        \begin{subfigure}[t]{\linewidth}
            \centering
            \strut\par
            \includegraphics[width=\linewidth]{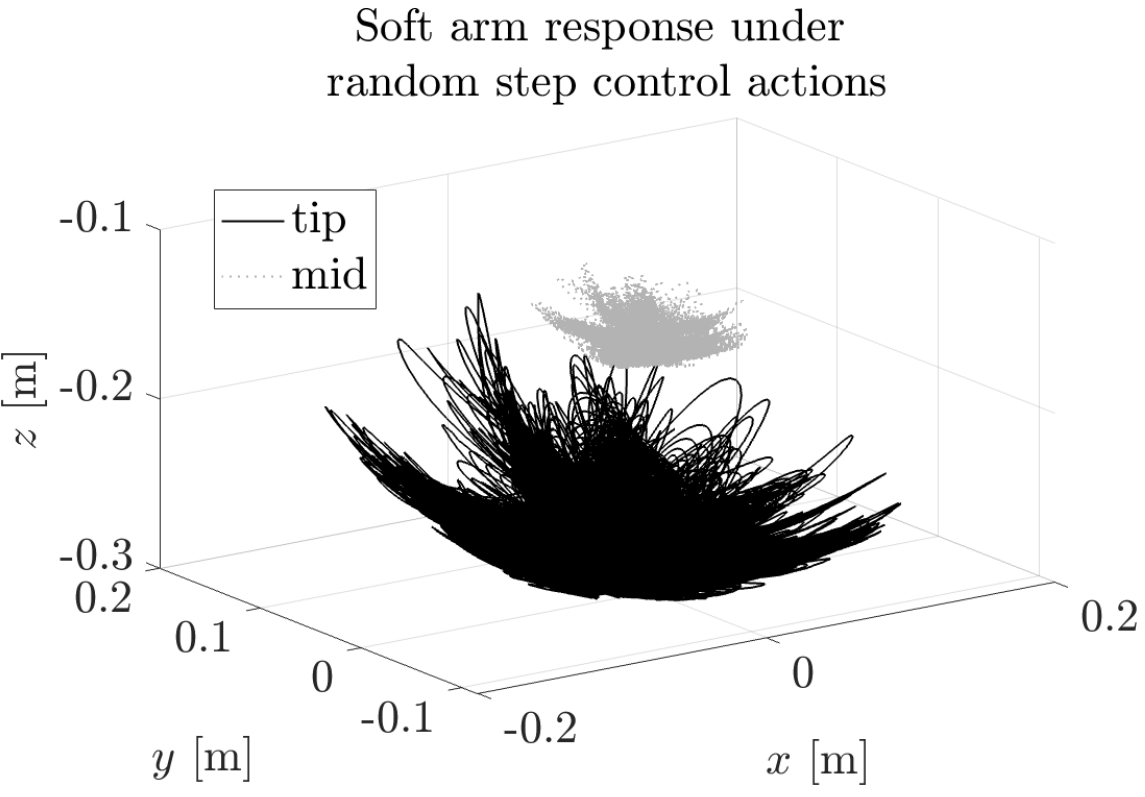}
            \setcounter{subfigure}{0}
            \caption{}
            \label{fig:step_response}
        \end{subfigure}
        \vspace{1pt}
        \begin{subfigure}[t]{\linewidth}
            \centering
            \strut\par
            \includegraphics[width=\linewidth]{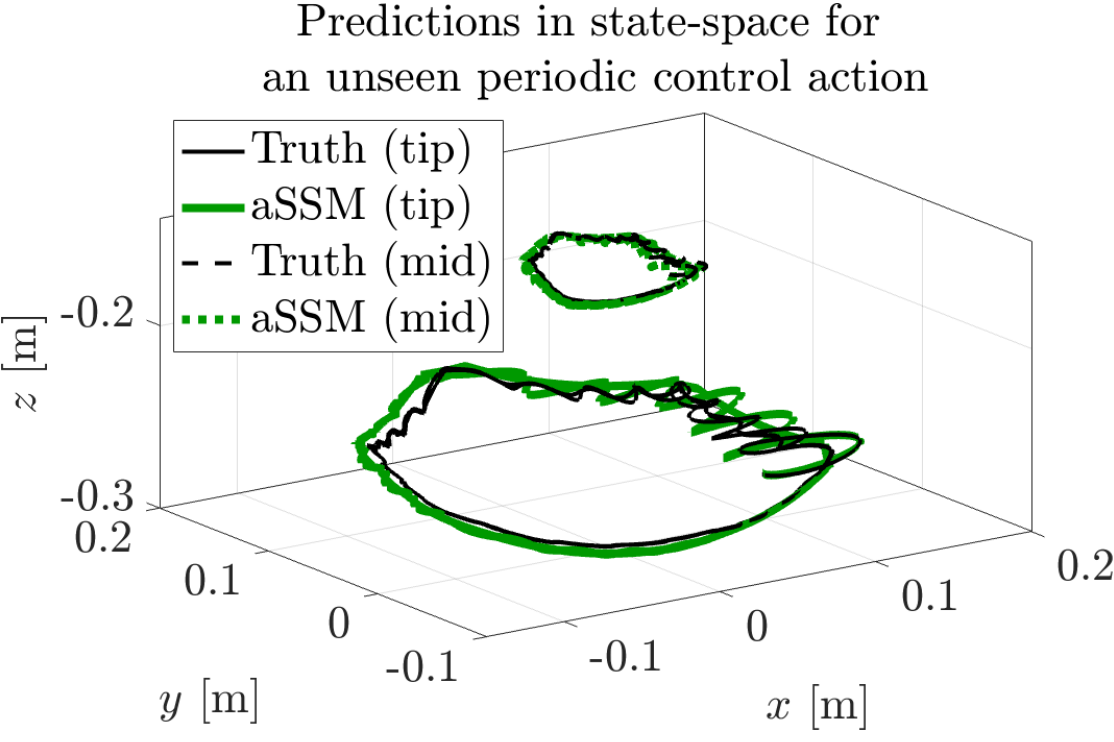}
            \setcounter{subfigure}{2}
            \caption{}
            \label{fig:periodic_state_space}
        \end{subfigure}
    \end{subfigure}\hfill
    % Column 2: (b) step response predictions, nudged down, beside the stack
    \begin{subfigure}[t]{0.535\textwidth}
        \centering
        \strut\par
        \vspace{14pt}
        \includegraphics[width=\linewidth]{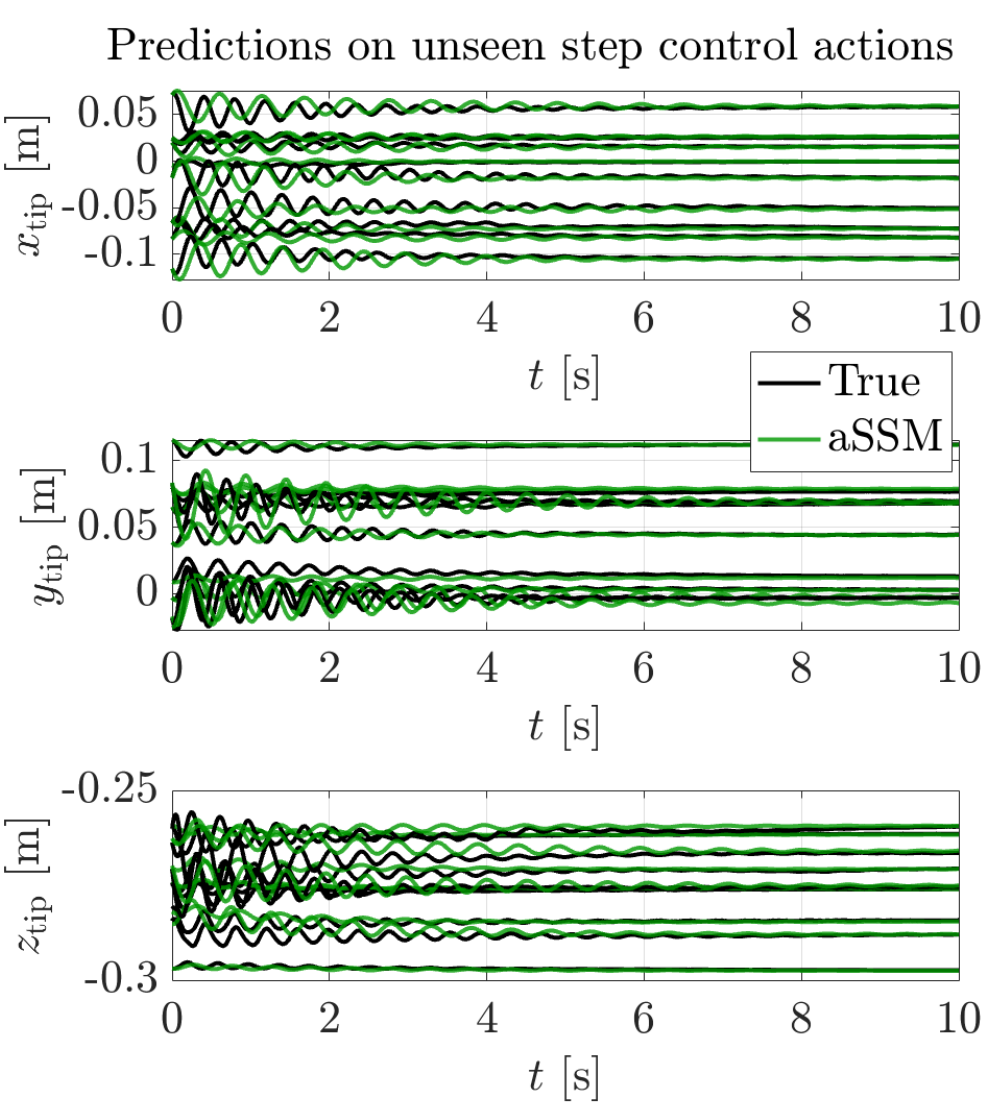}
        \setcounter{subfigure}{1}
        \caption{}
        \label{fig:pred_assm}
    \end{subfigure}
    \caption{\textbf{[Modeling SoPrA via an aSSM-reduced model]}. (a) Response of the SoPrA arm plotted in its state-space subjected to random step control actions. (b) Time series plots of the aSSM model predictions (green) and the SoPrA arm's true response (black) for unseen step control actions in the workspace of the SoPrA arm. (c) State-space plots of the aSSM model's predictions (green) and the SoPrA arm's true response (black) for a periodic control action.}
    \label{fig:sopra_validation_abc}
\end{figure}

We investigate the presence of normally attracting aSSMs in the observable space $\boldsymbol{s}$ of the SoPrA arm by commanding random step control action sweeps to it. These sweeps suffice to probe the leading-order aSSM geometry and its internal dynamics. As shown in Fig.~\ref{fig:step_response}, based on data collected from a 2.2-hour step sweep of 400 randomly sampled unique static control actions held for 20 seconds each, the SoPrA arm undergoes underdamped decays that eventually converge to unique static configurations in the observable space. Both of these characteristics support the assumption that normally attracting aSSMs exist in the phase space of the SoPrA arm (see Section \ref{sec:methods}). 

Using a spectrogram analysis of each decaying signal, we infer from the collected data, that the minimal dimension of the normally attracting fibers of the aSSM is $d=6$. This can also be deduced from the geometry of the SoPrA arm, which is symmetric in the $x-y$ plane neat the rest state. This implies that the modes of vibration along these axes are 1-1 resonant, which calls for the inclusion of at least 2 modes in the aSSM construction. The arm also admits longitudinal vibrations in the $z$ direction, which constitute another mode of vibration, thus explaining the $3$ dominant frequencies observed in the spectrogram analysis. All these minimal sets of modes are underdamped for the SoPrA arm, which is qualitatively different from the overdamped dynamics of tendon-driven trunk helicoid robots (see \citet{kaundinya2026precision}).

Our next step is to use the collected step control data and the inferred dimension $d=6$ of the aSSM fibers to learn the aSSM-reduced model for the SoPrA arm. That consists of three learnable functions: the critical manifold parametrization, which maps static control actions to static states; the aSSM parametrization; and the aSSM-reduced dynamics. The critical manifold parametrization is learned from the static states and control actions obtained by postprocessing the step control data in the observable space $\boldsymbol{s}$. The aSSM parametrization and dynamics cannot be learned directly in $\boldsymbol{s}$, as velocity information is required to model the dynamics, and an ambient space is needed to guarantee the embedding of the $d$-dimensional fibers of the aSSM geometry. To address these requirements, we employ delay embedding. Specifically, we construct a delay-embedded observable space by stacking the current state with its two time-delayed copies, in which a 6D fiber of the aSSM geometry can be embedded (see \citet{takens81}). We then use the collected step control data to learn all the aSSM-related functions in this delay-embedded observable space, following the procedure outlined in Section \ref{sec:assm_theory}.

The resulting aSSM-reduced model, which we will term the aSSM model, is of the form
\begin{align}
\label{eq:aSSM_leading_RL_main}
\mathbf{r} &= \mathbf{V}^{\mathrm T}(\boldsymbol{a}(t))\, \boldsymbol{\mu}\big(\boldsymbol{s} - \bar{\boldsymbol{s}}(\boldsymbol{a}(t))\big), \nonumber \\
\dot{\mathbf{r}} &= \mathbf{R}\big(\mathbf{r}, \bar{\boldsymbol{s}}(\boldsymbol{a}(t))\big), \\
\boldsymbol{s} &= \mathbf{W}\big(\mathbf{r}, \bar{\boldsymbol{s}}(\boldsymbol{a}(t))\big) + \bar{\boldsymbol{s}}(\boldsymbol{a}(t)), \nonumber
\end{align}
where $\mathbf{r} \in \mathbb{R}^6$ is reduced coordinates on the aSSM; $\mathbf{V}(\cdot)$ is the chart map of the aSSM; $\bar{\boldsymbol{s}}(\cdot)$ is the critical manifold parametrziation; $\boldsymbol{\mu}$ is a function performing delay embedding; $\mathbf{R}(\cdot)$ is the aSSM reduced dynamics; and $\mathbf{W}(\cdot)$ is the aSSM parametrization. We have found that quadratic polynomial approximations for $\mathbf{W}(\cdot)$, and $\mathbf{R}(\cdot)$, and cubic polynomial approximations for  $\bar{\boldsymbol{s}}(\cdot)$ suffice to model the dynamics of the SoPrA arm accurately. 

We plot the aSSM model's predictions against the SoPrA arm's true response on previously unseen step control actions in Fig.~\ref{fig:pred_assm} and on a previously unseen periodic control action sequence in Fig.~\ref{fig:periodic_state_space}. We observe that aSSM predictions qualitatively capture the decay profiles of all observable states. The normalized mean trajectory error achieved by the aSSM model on these test data was $9\%$ for step control data and $3 \%$ on the periodic control action data. We note that we do not have exact matches for any of the cases, and we attribute this to a hardware inaccuracy in the pressure actuation system. To illustrate this inaccuracy, in \edref{edfig:sopra_pressure}a, we plot the commanded pressure action variable $\boldsymbol{a}_2$ and compare it with the actual measured pressure in the chamber. We observe that, at all times, the pressure valves keep fluctuating and exhibit lags that lead to minor fluctuations in the arm's response (see \edref{edfig:sopra_pressure}b). This hardware issue is common in pressure control systems that are operated in a more dynamic regime. A mismatch between the commanded and measured pressure will result in accuracy issues for modeling and control in the subsequent sections. We aim to report results despite these limitations and test closed-loop control performance.

\subsection{Predictive power of models on unseen control actions across the full action space}\label{sec:black-box_model_comparisons}

As a comparison, we also learn the dynamics of the SoPrA arm by fitting neural-network-based models to trajectory data. This is done by approximating the state evolution as an autoregressive neural network $(b\text{-}m)$, defined as a map $\mathbf{f}$, 
\begin{equation}
\label{eq:ar_model}
\left(\boldsymbol{s}_{k+1}, \dots, \boldsymbol{s}_{k+m}\right) = \mathbf{f}\left(\boldsymbol{s}_k, \boldsymbol{s}_{k-1}, \dots, \boldsymbol{s}_{k-b+1},\; \boldsymbol{a}_k, \boldsymbol{a}_{k+1}, \dots, \boldsymbol{a}_{k+m-1}\right),
\end{equation}
where $b$ represents the memory window, and $m$ the forecasting window. The training data used to fit the neural networks is generated by issuing random-walk-like control actions that uniformly explore the state space. Importantly, to ensure a fair comparison, we keep the total amount of time-series training data collected for the neural network consistent with that collected for the aSSM model in Section~\ref{sec:softarm}. The exact training details for all neural network-based models on this dataset are provided in the Supplementary Information. We learn four neural-network-based models: two multilayer perceptrons (MLPs), one with $(b\text{-}m) \equiv (3\text{-}1)$ and the other with $(b\text{-}m) \equiv (3\text{-}8)$, a recurrent neural network (RNN), and a Transformer both with $(b\text{-}m) \equiv (32\text{-}8)$ and trained using state-of-the-art autoregressive training methods from \citet{li2025robotic}.

\begin{figure}[t]
        \includegraphics[width=\linewidth]{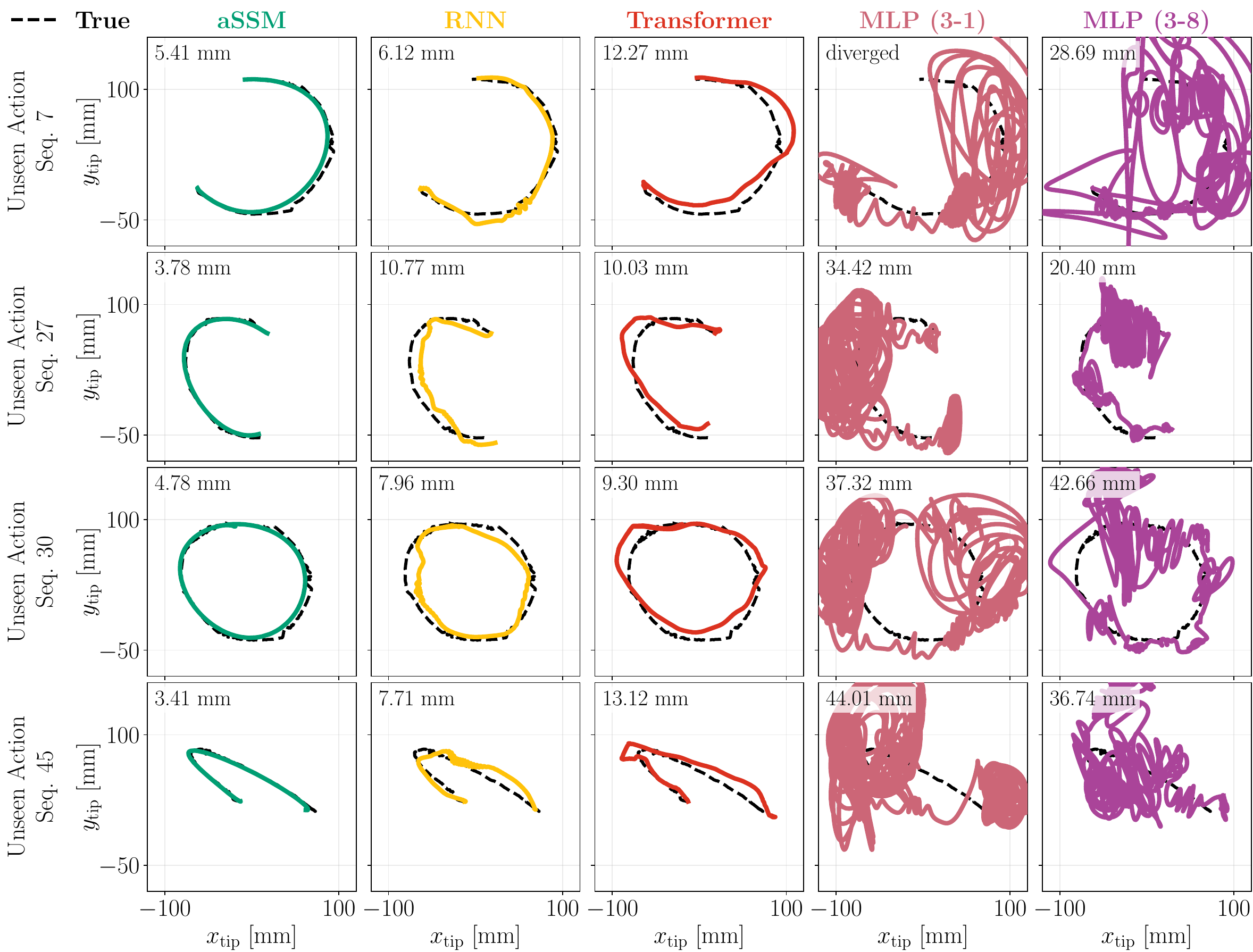}  
    \caption{\textbf{[Prediction results comparing the aSSM-reduced model against neural-network-based dynamics models]}. Plots in the $x_{\mathrm{tip}}-y_{\mathrm{tip}}$ plane for model predictions on four randomly sampled continuous control action sequences compared against the true response of the SoPrA arm (black). The mean error [mm] of the model prediction is reported in the top-left corner of each plot.}
    \label{fig:model_predictive_comparisons}
\end{figure}

We test the predictive power of all these models, including our aSSM model, on $50$ previously unseen continuous random control action sequences, uniformly sampled from the entire control action space. We pass these control action sequences to the SoPrA arm to generate the true responses, and we plot this with the model predictions in the $x_{\mathrm{tip}}-y_{\mathrm{tip}}$ plane. We present these results in Fig.~\ref{fig:model_predictive_comparisons} for four randomly chosen control action sequence runs. We observe that the aSSM model across these sets has the lowest mean error, and its predictions lie close to the true response of the arm.

The RNN and Transformer perform decently, qualitatively tracking the target, but remain inaccurate. In particular, the RNN tracks the true response reasonably well in the $x_{\mathrm{tip}}> 0 $ region but loses accuracy in other regions. In contrast, the Transformer uniformly exhibits larger errors throughout the workspace. On the other hand, both vanilla MLPs with simpler autoregressive network structures perform extremely poorly, with the error even diverging on some runs. Increasing the memory window from $b=1$ to $b=8$ occasionally reduces the error but also suffers from the same instabilities for other random runs. None of the autoregressive models provides a consistent and accurate prediction, which is not surprising, as the SoPrA arm does not exhibit significant memory effects due to material hysteresis or actuation delays in the collected data. We provide a detailed comparison and reasoning for these model behaviors in Section \ref{sec:pareto-doms}.

\subsection{aSSM-based RL policy for the soft arm}\label{sec:aSSM-RL}
We use the learned aSSM model from Section \ref{sec:softarm} to learn a reinforcement learning (RL) policy for tracking 2D random target tracks $\boldsymbol{\Gamma(t)} = \{\boldsymbol{\Gamma}_k\}_{k=1}^{T}$ in the $x_{\mathrm{tip}}-y_{\mathrm{tip}}$ plane of the SoPrA arm. The RL policy training proceeds by learning a conditional probability distribution over the space of actions, $\boldsymbol{\pi}_{\theta}(\boldsymbol{a}, \cdot)$, as a neural network.

A control action $\boldsymbol{a}_k$ at time step $k$ is sampled from $\boldsymbol{\pi}_{\theta}$ given an observation vector that depends on the current state $\mathbf{s}_{k}$, the previous state $\mathbf{s}_{k-1}$, the previously sampled control action $\boldsymbol{a}_{k-1}$, and the next target state $\boldsymbol{\Gamma}_{k+1}$. This sampled control action is passed to the aSSM model, which predicts the next state $\mathbf{s}_{k+1}$. A reward $\rho_k = \rho(\mathbf{s}_{k+1},\boldsymbol{a}_k,\boldsymbol{\Gamma}_{k+1})$ is evaluated based on these predictions, the control action that generated it, and the target state. A cumulative reward sum, called the \textit{episode return}, is computed after rolling out the model and policy across the target track horizon. 

Following standard offline model-based RL techniques, we launch and compute episode returns in parallel across a large set of initial conditions sampled from the training data used to fit these models. This procedure is referred to as \textit{imagination} in the RL community. The weights $\theta$ of $\boldsymbol{\pi}_{\theta}$ are updated to maximize the mean episode return over an imagination cycle, and each update is defined as a training iteration of the RL policy. In Section \ref{sec:rl}, we provide details on the problem formulation and the RL training algorithm, and in the Supplementary Information, we provide the reward function, the hyperparameters, and the number of initial conditions launched in one imagination cycle.

We perform the same RL training procedure using only the RNN and Transformer models discussed in Section~\ref{sec:black-box_model_comparisons}, since these models provided stable predictions. In \edref{edfig:rl_reward}, we plot the mean imagination episode return for increasing training iterations to validate if the policy updates indeed converge by maximizing the reward. We plot these for three policies: aSSM-RL, a policy trained on an aSSM model; RNN-RL, a policy trained on an RNN model; and Transformer-RL, a policy trained on a Transformer model. We observe that aSSM-RL smoothly and rapidly reaches a good mean imagination episode return value within $200$ iterations, and steadily improves as the training iterations increase. Both the RNN-RL and Transformer-RL only reach a decent mean imagination episode return value after $800$ training iterations. Compared to aSSM-RL, training neural network model-based policies often stagnates, and improvements come as short performance gains. They also show erratic behavior as the hyperparameter is swept (shaded region in \edref{edfig:rl_reward}), signaling the black-box nature of these models.

We perform a total of $5,000$ training iterations, which ensures convergence of all policies. We note that the successful performance of final converged policies on the model does not imply the same on the hardware SoPrA arm. We follow up on this point with a discussion in Section \ref{sec:pareto-doms} that systematically compares various features of these models for hardware deployment. 

\subsection{Pareto dominance in model complexity, learning time, and accuracy}\label{sec:pareto-doms}

We compare here the data-driven models and the RL policies trained on them in terms of model complexity, computation time, and ability to accurately capture hardware dynamics and performance accuracy in closed-loop on hardware. 

\begin{figure}[t]
    \centering
    \begin{subfigure}[t]{\linewidth}
        \centering
        \includegraphics[width=0.8\linewidth]{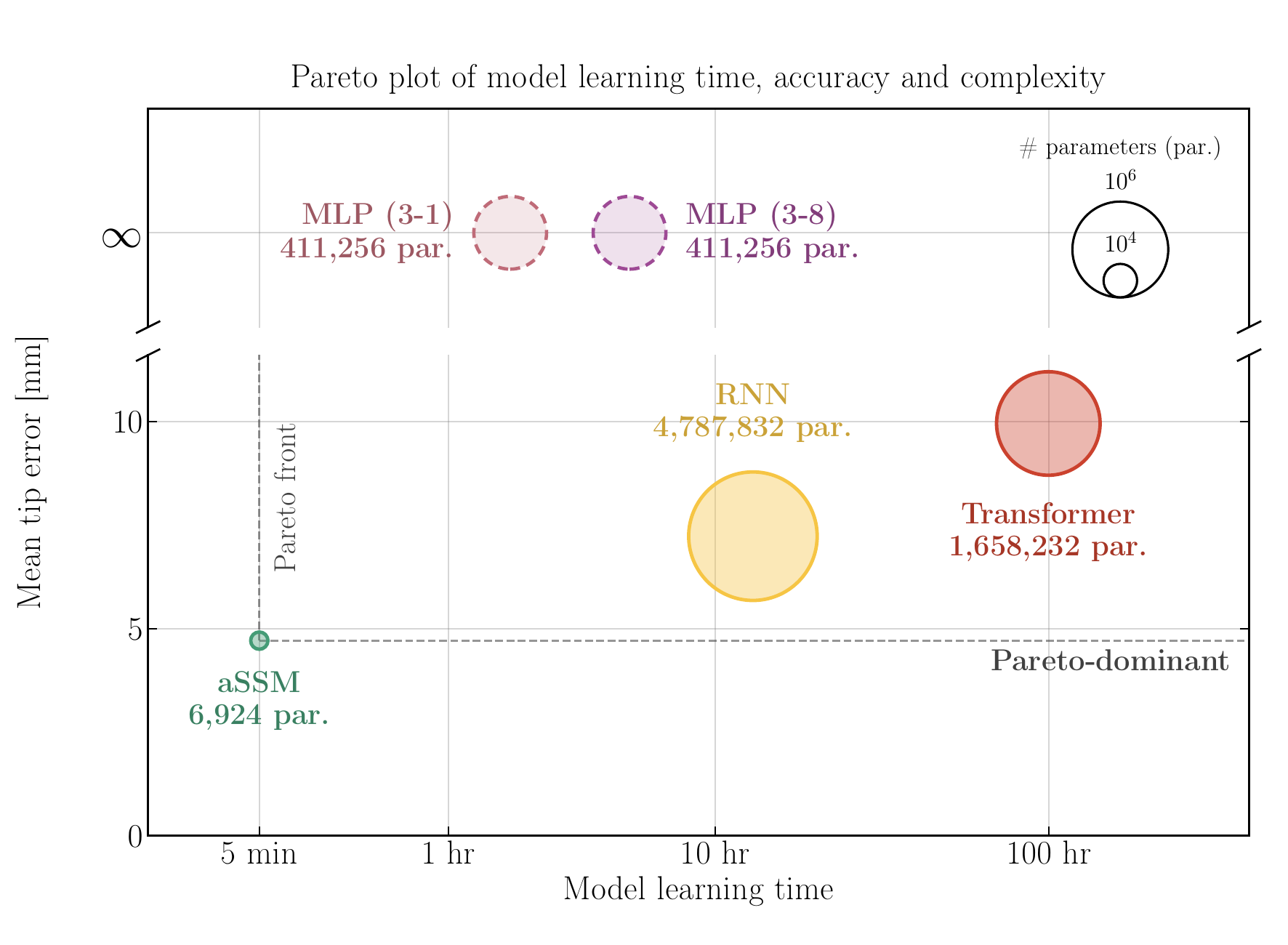}
        \caption{}
        \label{fig:pareto_model}
    \end{subfigure}

    \vspace{6pt}

    \begin{subfigure}[t]{\linewidth}
        \centering
        \includegraphics[width=0.8\linewidth]{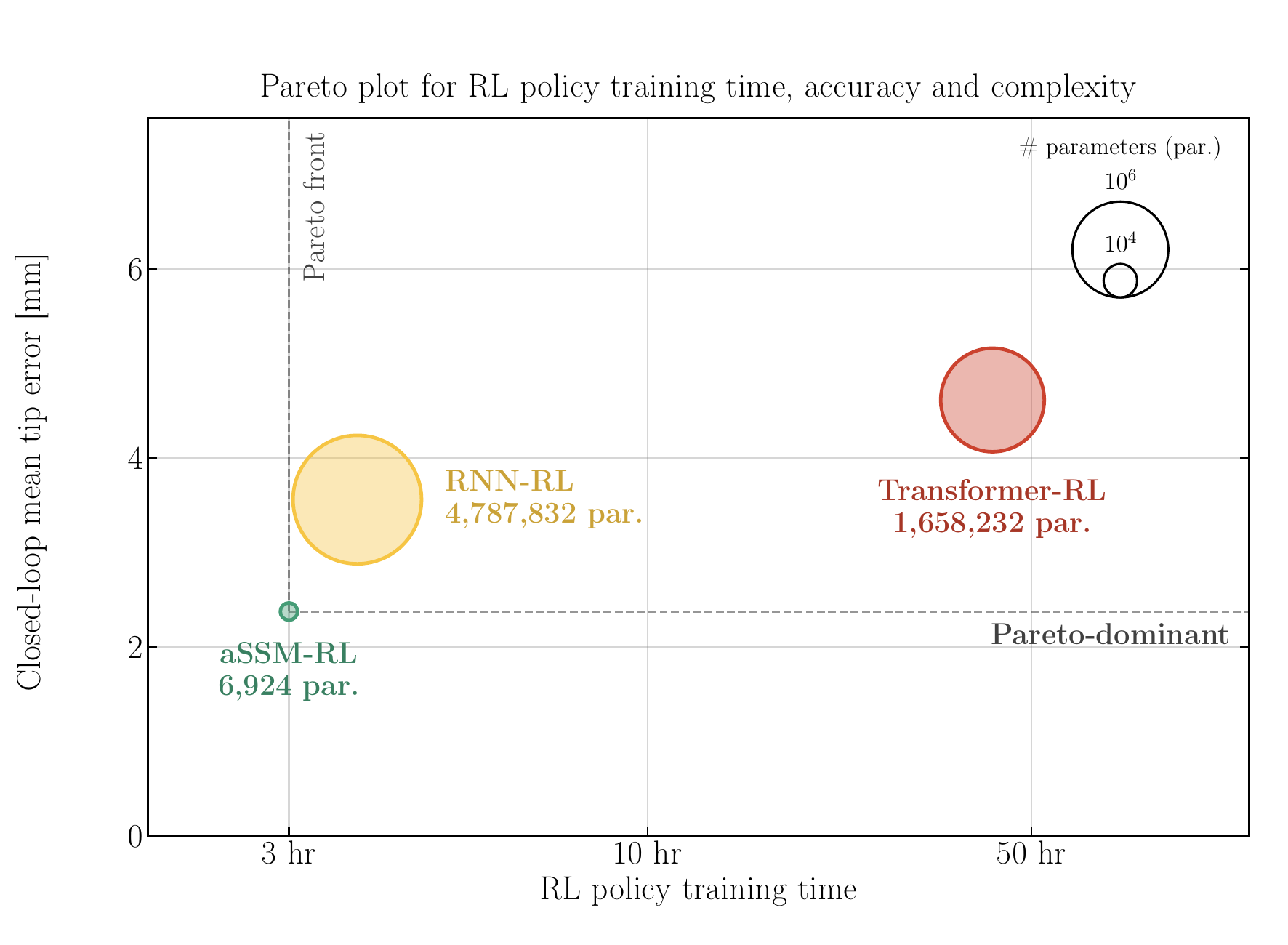}
        \caption{}
        \label{fig:pareto_RL}
    \end{subfigure}
    \caption{\textbf{[Pareto plot comparisons for learned models and policies]}. (a) Mean prediction error of the SoPrA arm's tip position for the random unseen control action experiments in Section \ref{sec:black-box_model_comparisons}, plotted against model learning time, with bubble size denoting the number of model parameters. The aSSM model is Pareto dominant across these metrics. (b) Mean prediction error of the SoPrA arm's tip position across the closed-loop runs in Section \ref{sec:closedloop}, plotted against policy training time, with bubble size denoting the number of model parameters. The aSSM-RL policy is Pareto dominant across these metrics.}
    \label{fig:pareto}
\end{figure}

Our first comparison is a Pareto plot in Fig.~\ref{fig:pareto_model} that focuses only on the performance of the dynamical models. We use the $50$ randomized control action experiments from Section \ref{sec:black-box_model_comparisons} to plot the mean prediction error of the models across these runs along the vertical axis in Fig.~\ref{fig:pareto_model}. The horizontal axis in this figure shows the time taken to learn the various models, with the training dataset size held constant. The number of model parameters is denoted by bubbles, with each bubble's radius indicating the size of the model. Across all metrics, the aSSM model is Pareto-dominant and defines the Pareto front. As a simple polynomial model (see Section~\ref{subsec:datadriven_RL} and eqs.~(\ref{eq:W_poly_RL})-(\ref{eq:R_poly_RL})), the aSSM model requires only $5$ minutes to learn, which is orders of magnitude lower than the learning time of the other models. The RNN model lies closest to the Pareto front, where it performs reasonably well in terms of predictive accuracy. However, it is ultimately too complex, requiring significantly longer learning times on the order of hours. The Transformer model is much further from the Pareto front. Despite having fewer parameters than the RNN model, its architecture requires a massive 100 hours of learning time. The instability of MLPs keeps them out of this plot. They are easier to learn because of their moderate sizes, but their erratic accuracy prevents their use for controls. Hence, we omit these models for learning control policies.

\FloatBarrier

For the trained model-based RL policies, we offer similar comparisons of the figure, in Fig.~\ref{fig:pareto_RL}, but now with the vertical axis denoting mean closed-loop prediction accuracy on hardware across the tasks presented in Section~\ref{sec:closedloop}. The horizontal axis of the figure denotes the time taken to train the RL policies. The aSSM-RL is still a Pareto-dominant model. In this case, the RNN-RL is close to the Pareto front but ultimately loses out on closed-loop accuracy, and the RNN model architecture is approximately 500 times more complex than the aSSM model, which adds to the computational burden of training the RL policy only. We note here that Transformer, though considered state-of-the-art, appears to perform the worst across all metrics. Overall, we attribute these trends in neural network models to the limited dataset they can access during training, a practical constraint imposed by hardware.

\subsection{Closed-loop target tracking results}\label{sec:closedloop}
We finally close the results section, with closed-loop experiments on the hardware SoPrA arm using our trained aSSM-RL policy. We set up two target shapes in the $x_{\mathrm{tip}}-y_{\mathrm{tip}}$ plane, the lemniscate resembling a Figure-8 pattern, and a square. The lemniscate has a maximum span of $100$ [mm] in length and $50$ [mm] in width, and the target track moves at a speed of $17.5$ [mm/s]. The square's side length is $100$ [mm]. Both of these targets span the maximal regions of the  SoPrA arm's workspace (see Fig.~\ref{fig:step_response}). For the square target, we additionally consider three cases varying in speed. The slowest has a speed of $27.5$ [mm/s], while the next two are 1.6 and 3.2 times faster. 

\begin{figure}[H]
    \centering
    \begin{subfigure}[t]{\linewidth}
        \centering
        \makebox[\linewidth][c]{%
            \includegraphics[width=\linewidth]{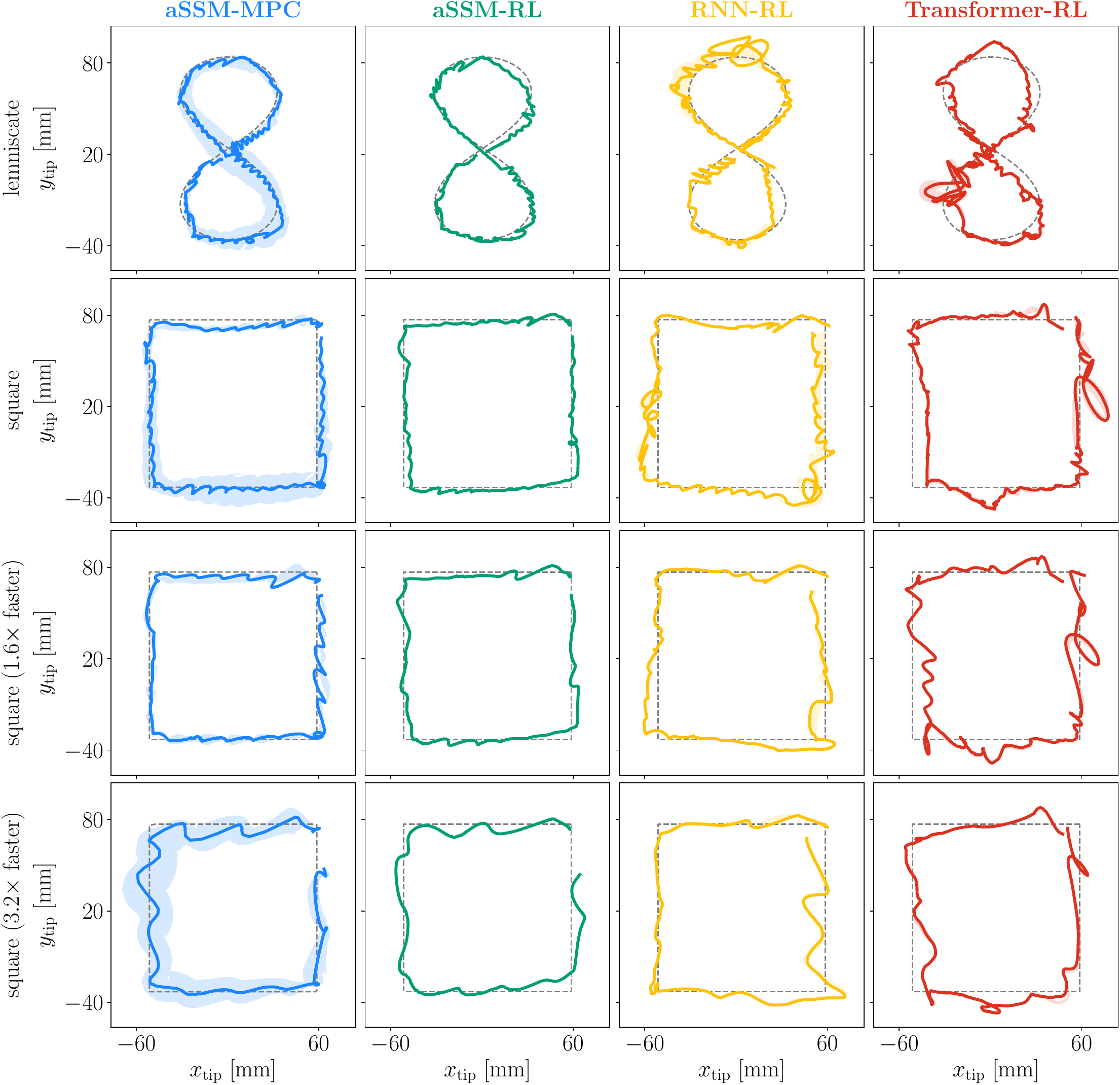}}
        \caption{}
        \label{fig:cloop}
    \end{subfigure}

    \vspace{6pt}

    \begin{subfigure}[t]{\linewidth}
        \centering
        \makebox[\linewidth][c]{%
            \includegraphics[width=\linewidth]{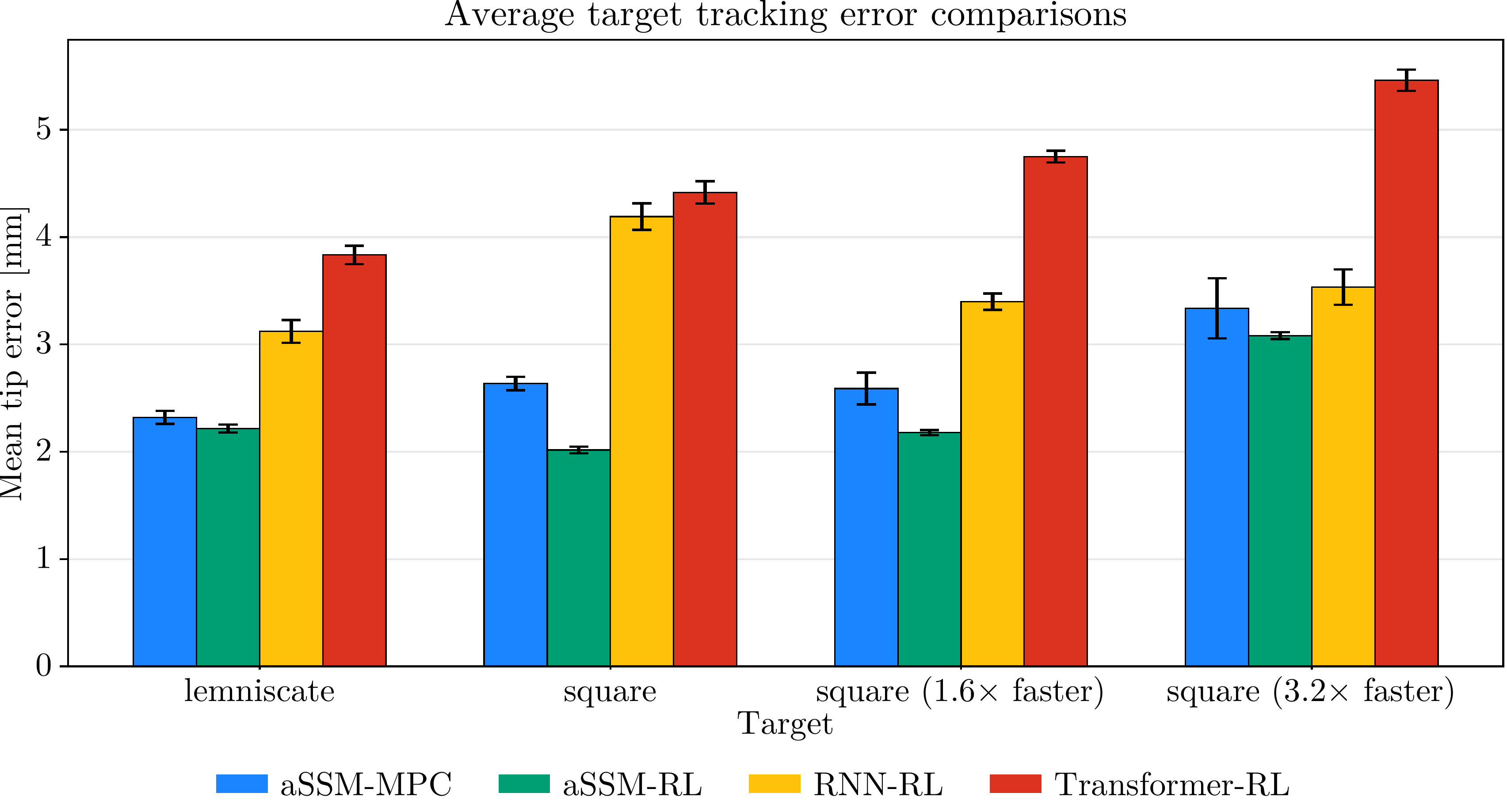}}
        \caption{}
        \label{fig:clooperror}
    \end{subfigure}
    \caption{\textbf{[Closed-loop experiments using learned model-based RL policies]}. (a) Closed-loop results for four target tracks for aSSM-MPC (blue), aSSM-RL (green), RNN-RL (yellow), and Transformer-RL (red). Solid lines show the mean over 10 repeated runs, and shaded regions show the deviation. (b) Average tracking error compared across all target tracks. The deviation bar indicates the performance variation across 10 repeated runs of the closed-loop experiments.}
    \label{fig:closedloop}
\end{figure}

In Fig.~\ref{fig:cloop}, we provide closed-loop results for these target tracks with comparisons against RNN-RL, Transformer-RL, and also aSSM-MPC, a model predictive controller (MPC) similar to the one used in \citet{kaundinya2026precision}, minimizing a quadratic cost to track the target. We quantitatively evaluate the closed-loop prediction error to the target and plot these as bar plots in Fig.~\ref{fig:clooperror}. We note that our error penalizes spatial track-following. If we also account for strict temporal accuracy, all closed-loop policy prediction errors increase by a factor of $2$ [mm] on the lemniscate and the slow square due to hardware latency (see the Supplementary Information). The aSSM-RL method achieves the best performance across various target tracks, with aSSM-MPC following it closely as expected. Both neural-network-based models trace out a track, but as the error increases, they become unstable, causing them to oscillate a bit and leading to larger errors. 

We recall that the validity of using the aSSM model for controls relies on the time-scale separation assumption. We can quantitatively verify the assumption on a target track by evaluating a non-dimensional slowness measure $r_s$, defined in eq.~(\ref{eq:slowness_RL}). The lemniscate and the slow-speed square have $r_s = 0.1$. The faster squares have an $r_s=0.3$ and $r_s=0.8$. We expect the theory to hold when $r_s\ll1$, and in practical problems, we have observed that even for $r_s < 1$ we obtain good closed-loop performance (see \citet{Kaundinya2026} and \citet{kaundinya2026precision}). This is indeed the case across the target tracks considered here, with performance degrading slightly as the target speed increases. The aSSM-MPC's limitation is apparent in the fastest target tracks, since it operates at 50 Hz due to the computational burden of solving the problem in real time. The aSSM-RL policy, on the other hand, operates at 100 Hz, which avoids the latency issue of MPC methods.

\begin{figure}[H]
    \makebox[\linewidth][c]{%
        \includegraphics[width=\linewidth]{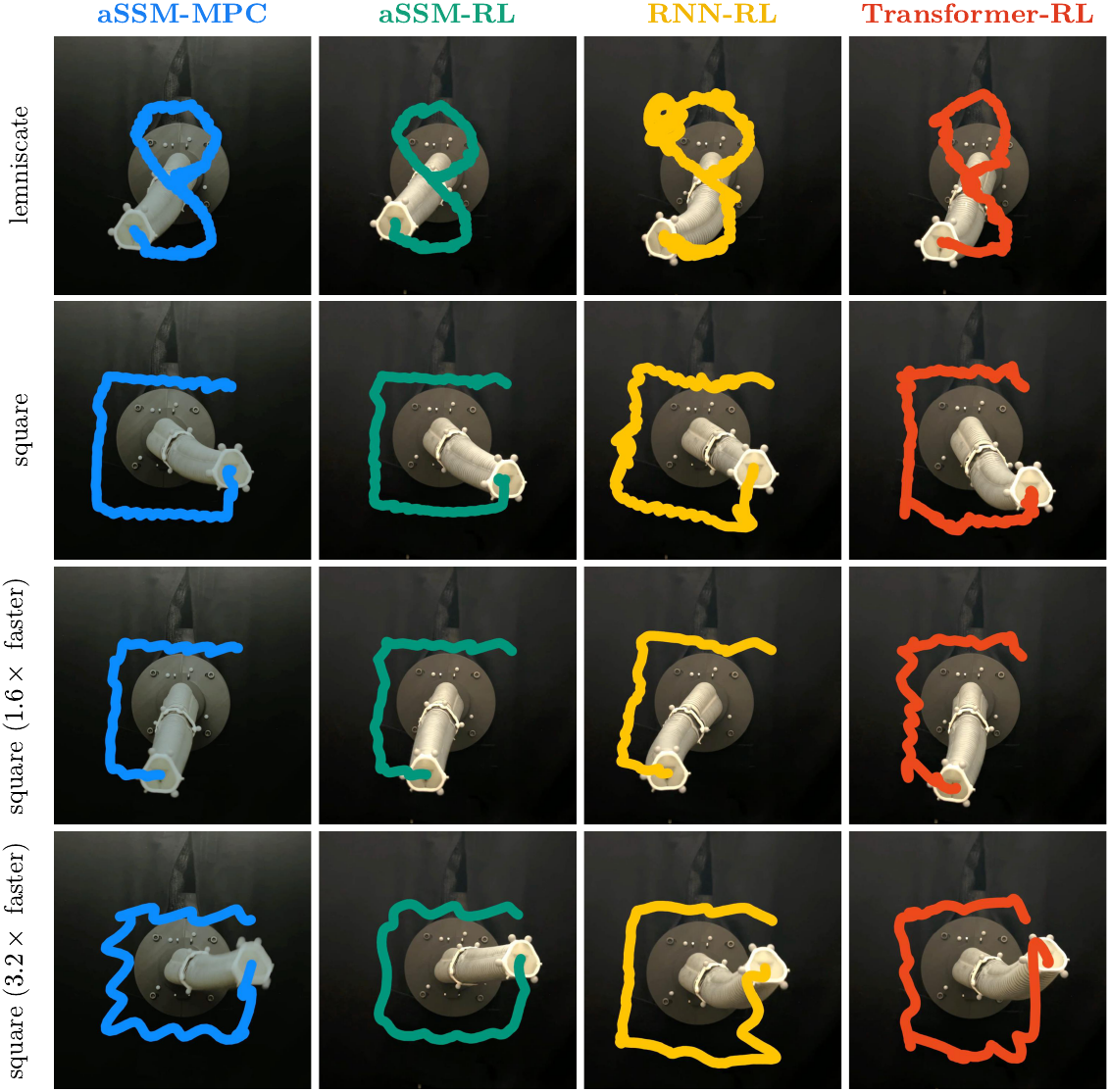}}
    \captionof{figure}{\textbf{[Closed-loop control of SoPrA]}. Snapshots of the SoPrA arm during the various closed-loop experiments for all the target tracks. Each snapshot comes from one run, randomly selected from the 10 repeated runs shown in Fig.~\ref{fig:cloop}. The traced path by the tip of the arm is plotted as a thick line for each prediction in its respective color.}
    \label{fig:snaps}
\end{figure}

We also provide snapshots of the SoPrA arm during closed-loop operation for all target tracks and across all policies in Fig.~\ref{fig:snaps}. We note that the aSSM-RL achieves simpler nonlinear shape profiles for the arm, ensuring easy tracking. In contrast, the RNN-RL and transformer-RL assume extreme bending shapes, which prevent continuous, consistent tracking of the target.

\subsection{Closed-loop results with potential hardware upgrades in the pressure-driven actuation system}\label{sec:upgrades_cl}

In Section \ref{sec:softarm}, in particular in \edref{edfig:sopra_pressure}a, we highlighted the hardware limitation that the controller-commanded pressure does not exactly match the measured pressure in the chambers. We expected this mismatch to cause accuracy issues during closed-loop experiments on hardware. Indeed, we observe that no policy prediction is spared, in Fig.~\ref{fig:cloop}, they appear as minor wiggles along the entire closed-loop prediction.

Upgrading to more expensive pressure valves is expected to improve accuracy by reducing the mismatch between commanded and measured pressure caused by the dynamic nature of the tasks. It will also allow the practitioner to test their methods on tasks involving faster target tracks. To aid the practitioner in deciding on the upgrade, from the mismatch data in \edref{edfig:sopra_pressure}a, we evaluated numerically that a moving-average procedure with a window of 0.3 seconds (the current time lag in the pressure valve between commanded and measured) removes most of the wiggles, thus mimicking closed-loop predictions when the mismatch is absent. We applied this filtering procedure to our policy prediction from Section \ref{sec:closedloop} and report these results in \edref{fig:cloop_filter}, and also provide revised mean error bar plots in \edref{fig:clooperror_filter}.

The results show promise that upgrading to better valves guarantees improved performance across all target tracks. Only large oscillations intrinsic to the SoPrA arm are visible, which the aSSM model accurately models and limits compared to the RNN and Transformer models. The error bars also point to an interesting trend. Our heuristic numerical filtering procedure reduces errors in the RNN-RL and Transformer-RL predictions to a greater extent than in the aSSM model, indicating that, despite these pressure mismatch issues, our aSSM-RL method explores the best possible policy to reduce the existing hardware mismatch error.

\section{Discussion}\label{sec:discussion}

We have demonstrated that hardware soft robots can be accurately controlled for dynamic target tracking via Reinforcement Learning (RL) control policies trained solely on a faithful, reduced dynamics model of the soft robot. This advance is made possible by the fact that the dominant dynamics of soft robots lie on a very low-dimensional adiabatic spectral submanifold (aSSM). Recent methods allow for the rapid extraction of reduced models based on aSSM theory from hardware data alone.  

For a pneumatically actuated SoPrA soft robotic arm, we have learned a 6D aSSM model from only partial observational data recorded by cameras (see Section~\ref{sec:softarm}). The learned 6D aSSM model delivers strong predictive capability and generalizes to time-varying control actions not present in the training dataset (see Section~\ref{sec:black-box_model_comparisons}). When compared with state-of-the-art autoregressive neural-network-based models, our aSSM model is Pareto dominant across prediction accuracy and model learning time, offering the fastest model to learn and the simplest with parameters in the thousands compared to the others in the millions (see Section~\ref{sec:pareto-doms}). This simplicity also allows the aSSM model to be used in classical model predictive control schemes for real-time control, which are otherwise intractable to use with large models (see aSSM-MPC results in Section \ref{sec:closedloop}).

RL control policies trained on our aSSM model have shown stable performance across a wide range of RL hyperparameters and converge rapidly in a couple of hours, unlike policies trained on autoregressive neural-network-based models such as the RNN and the Transformer (see Section~\ref{sec:aSSM-RL}). These RL policies were trained without interacting with the hardware soft robot and only explore the control action space using models learned on two hours of hardware data. By machine-learning standards, this is a small dataset, so one would expect the resulting RL policies to transfer poorly to the real world.

Our closed-loop results in Section \ref{sec:closedloop} support this expectation: the RNN-RL and Transform-RL policies produce large errors when tracking targets in real time. In contrast, the aSSM-RL policy provides consistent, accurate closed-loop performance. This is because, the aSSM provides a faithful reduced-order representation of the robot’s dynamics and therefore requires only data capturing the dominant behaviors that are ubiquitous and thus do not demand large volumes of data, unlike black-box approaches.

Our discovery of reliable, data-efficient aSSM-RL policies for soft robots opens exciting avenues for applying these techniques to challenging, application-oriented problems in interactive environments, such as surgical assistance or underwater exploration. Learning aSSM-RL policies for soft robotic hands performing dynamic dexterous manipulation is already within reach and currently under exploration. A longer-term goal is to learn aSSM models from video alone, without physical markers. Policies designed using such models will allow us to control complex soft robot morphologies and operate them in unstructured environments.

\section{Methods}\label{sec:methods}

\subsection{Adiabatic Spectral Submanifolds: Theory and Data-Driven Methodology}\label{sec:assm_theory}

\subsection*{Theory}\label{subsec:theory_RL}

The governing equations of soft robots can be written abstractly as a first-order dynamical system
\begin{equation}
\label{eq:eom_RL}
\dot{\mathbf{x}} = \mathbf{F}(\mathbf{x}, \boldsymbol{a}(t)), \qquad \mathbf{x} \in \mathcal{X} \subseteq \mathbb{R}^{n}, \quad \boldsymbol{a} \in \mathcal{C} \subseteq \mathbb{R}^{n_a},
\end{equation}
with $n$ typically very large due to the continuum nature of the problem. Here $\mathbf{x}$ is the full phase space of the robot, $\boldsymbol{a}$ is the control action drawn from the admissible set $\mathcal{C}$, and $\mathbf{F}$ is a smooth nonlinear function that captures the geometric and material deformations of the arm.

For a static control input $\bar{\boldsymbol{a}} \in \mathcal{C}$, the robot deforms into a static configuration. Assuming that the collection of all these static configurations are attracting across $\mathcal{C}$, they form a normally attracting critical manifold (see \citet{fenichel73}), 
\begin{equation}
\label{eq:L0_RL}
\mathcal{L}_0 = \big\{\, (\mathbf{x},\bar{\boldsymbol{a}}) \in \mathcal{X} \times \mathcal{C} \ : \ \mathbf{x} = \bar{\mathbf{x}}(\bar{\boldsymbol{a}}) \,\big\}.
\end{equation}
If we further assume the linearization about each static configuration has a strict spectral splitting after $d$ slowest eigenvalues, uniformly over $\mathcal{C}$,
\begin{equation}
\label{eq:spec_RL}
\mathrm{Re}(\lambda_n(\bar{\boldsymbol{a}})) \leq \dots \leq \mathrm{Re}(\lambda_{d+1}(\bar{\boldsymbol{a}})) < \mathrm{Re}(\lambda_{d}(\bar{\boldsymbol{a}})) \leq \dots \leq \mathrm{Re}(\lambda_{1}(\bar{\boldsymbol{a}})) < 0,
\end{equation}
we can define the slow spectral subbundle $\mathbf{E}(\bar{\boldsymbol{a}})$ associated with the eigenspaces of the $d$ slowest eigenvalues. Under further non-resonance conditions, this subbundle of the linearized problem survives as a slow spectral submanifold bundle
\begin{equation}
\label{eq:A0_RL}
\mathcal{A}_0 = \mathcal{W}\big(\mathbf{E}(\bar{\boldsymbol{a}})\big),
\end{equation}
which is also normally attracting and anchored to the critical manifold $\mathcal{L}_0$ (see \citet{haller24_wa}). The slow spectral submanifold bundle $\mathcal{A}_0$ captures the dominant dynamics of system eq.~(\ref{eq:eom_RL}) with static control actions. 

\citet{haller24_wa} and \citet{Kaundinya2026} show that both $\mathcal{L}_0$ and $\mathcal{A}_0$ survive into nearby diffeomorphic invariant manifolds, $\mathcal{L}_{\epsilon}$ the slow manifold and $\mathcal{A}_\epsilon$ the adiabatic spectral sumanifold (aSSM) when the control action is of the form 
\begin{equation}
\label{eq:control_action_split}
\boldsymbol{a}(t) = \boldsymbol{a}^s(\epsilon t) + \boldsymbol{a}^d(t), 
\end{equation}
with $\|\boldsymbol{a}^d(t)\| < \Delta$, and  $0 < \epsilon \ll 1$. General control actions in control operations admit this splitting, in which $\boldsymbol{a}^s(t)$ is relatively slowly varying but can be large compared to a small, uniformly bounded deviation control action $\boldsymbol{a}^d(t)$. We refer to \citet{haller24_wa} for details of the aSSM-reduced model derived from invariance equations and to \citet{Kaundinya2026} for details in the control setting.

On hardware, the full phase space $\mathbf{x}$ is only partially observable: cameras record only a sparse set of positional data $\boldsymbol{s} \in \mathcal{S} \subseteq \mathbb{R}^{n_s}$ with $n_s \ll n_x$. In that case, the critical manifold definition carries over to the observable state space as $\bar{\boldsymbol{s}}(\bar{\boldsymbol{a}})$. 
The state $\boldsymbol{s}$ has no velocity information and hence does not guarantee embedding the dynamics of the $ d$-dimensional fibers of the frozen aSSM geometry $\mathcal{A}_0$. To successfully embed the aSSM and its dynamics in an ambient space, by Takens' theorem (see \citet{takens81}),  we need to delay embed the positional observable $\boldsymbol{s}$, at least  $p \geq \Big\lfloor \frac{2d+1}{n_y} \Big\rfloor$ times.

In this delay-embedded observable space, assuming the control action satisfies eq.~(\ref{eq:control_action_split}), we denote  $\mathbf{V}(\boldsymbol{a}(t)) \in \mathbb{R}^{(p+1)n_s \times d}$ whose columns span the time-varying slow spectral subspace $\mathbf{E}(\boldsymbol{a}(t))$ along the trajectory $\boldsymbol{\mu}(\boldsymbol{\bar{s}}(\boldsymbol{a}(t)))$, and $\mathbf{r} \in \mathbb{R}^{d}$ the reduced coordinate along this subbundle, the leading-order aSSM-reduced model takes the form
\begin{align}
\label{eq:aSSM_leading_RL}
\mathbf{r} &= \mathbf{V}^{\mathrm T}(\boldsymbol{a}(t))\, \boldsymbol{\mu}\big(\boldsymbol{s} - \bar{\boldsymbol{s}}(\boldsymbol{a}(t))\big), \nonumber \\
\dot{\mathbf{r}} &= \mathbf{R}\big(\mathbf{r}, \bar{\boldsymbol{s}}(\boldsymbol{a}(t))\big), \\
\boldsymbol{s} &= \mathbf{W}\big(\mathbf{r}, \bar{\boldsymbol{s}}(\boldsymbol{a}(t))\big) + \bar{\boldsymbol{s}}(\boldsymbol{a}(t)), \nonumber
\end{align}
where $\mathbf{W}(\cdot)$ is the aSSM parametrization and $\mathbf{R}(\cdot)$ is the aSSM reduced dynamics. 

\begin{rem}
\textbf{[Discrete-time aSSM-reduced model for reinforcement learning (RL)]}
Most model-based RL algorithms require a discrete dynamics model. We obtain the discrete reduced model by integrating the continuous leading-order dynamics (\ref{eq:aSSM_leading_RL}) over a time step  $\Delta t$ under a zero-order hold, i.e., the action is held fixed at $\boldsymbol{a}_k$ over $[t_k, t_k + \Delta t)$. This results in a discrete map of the aSSM-reduced model, 
\begin{align}
\label{eq:aSSM_discrete_RL}
\mathbf{r}_k &= \mathbf{V}^{\mathrm T}(\boldsymbol{a}_k)\, \boldsymbol{\mu}\big(\boldsymbol{s}_k - \bar{\boldsymbol{s}}(\boldsymbol{a}_k)\big), \nonumber \\
\mathbf{r}_{k+1} &= \mathbf{R}_d\big(\mathbf{R};\, \mathbf{r}_k, \bar{\boldsymbol{s}}(\boldsymbol{a}_k), \Delta t\big), \\
\boldsymbol{s}_{k+1} &= \mathbf{W}\big(\mathbf{r}_{k+1}, \bar{\boldsymbol{s}}(\boldsymbol{a}_k)\big) + \bar{\boldsymbol{s}}(\boldsymbol{a}_k), \nonumber
\end{align}
 where $\mathbf{R}_d(\cdot)$ is the discrete reduced dynamics map approximated via numerical methods from the continuous aSSM reduced dynamics $\mathbf{R}(\cdot)$.
\end{rem}

\begin{rem}
\textbf{[Higher-order correction to the aSSM-reduced model]}
\citet{kaundinya2026precision} provides higher-order corrections to the aSSM-reduced dynamics, which depend on the control action rate. Based on the assumptions of the control action splitting in eq.~(\ref{eq:control_action_split}), the higher-order corrected aSSM reduced model is of the form
\begin{equation}
\label{eq:aSSM_sub_RL}
\dot{\mathbf{r}} = \mathbf{R}\big(\mathbf{r}, \bar{\boldsymbol{s}}(\boldsymbol{a}(t))\big) + \mathbf{B}(\boldsymbol{a}(t))\, \mathbf{D}_{\mathbf a}\bar{\boldsymbol{s}}(\boldsymbol{a}(t))\, \dot{\boldsymbol{a}} + O\!\left((\epsilon+\Delta)^2,\ \epsilon|\mathbf{r}|,\ \Delta|\mathbf{r}|\right),
\end{equation}
with $\mathbf{B}(\boldsymbol{a}(t))$ a calibration matrix learned from the leading-order aSSM-reduced model. We ignore this correction here, since for our pneumatically actuated arm, the inconsistencies between the commanded and measured pressures in the actuation system make it impossible to learn an accurate calibration matrix.
\end{rem}

\begin{rem}
\textbf{[Time-scale separation for target tracking]} Our aSSM-reduced models are valid only under the time-scale separation assumption stated in Section~\ref{sec:softarm}. We quantify this with the non-dimensional slowness measure $r_s$ for practical application,
\begin{equation}
\label{eq:slowness_RL}
r_s = \frac{\displaystyle\frac{1}{t_f-t_i}\int_{t_i}^{t_f}\big\|\dot{\boldsymbol{\Gamma}}(\tau)\big\|\,d\tau}
{\overline{\displaystyle\frac{1}{t_d}\int_{0}^{t_d}\big\|\dot{\boldsymbol s}_{\text{decay}}(\tau)\big\|\,d\tau}},
\end{equation}
where $\boldsymbol s_{\text{decay}}$ is a decaying trajectory of the robot, $t_d$ is the duration of that decay, and the overbar denotes an average across the various decays obtained by passing step control actions spanning $\mathcal{C}$. The numerator is the mean speed along the target $\boldsymbol{\Gamma}(t)$ over its horizon $[t_i,t_f]$. The theory is guaranteed for $r_s \ll 1$, but our method performs well in practice up to $r_s \approx 1$ (see \citet{Kaundinya2026} and \citet{kaundinya2026precision}).
\end{rem}

\subsection*{Data-driven methodology}\label{subsec:datadriven_RL}

We collect the robot's response to step control actions in the observable space. The step control actions are generated by performing jumps, each held for $t_{\mathrm{decay}}$ seconds, between the $m$ most distant control actions for every element of the static action set, which we sample by Latin hypercube sampling (see \citet{latin_hypercube}). We set the total number of jumps, $m = 4$, for the step data collection for the SoPrA arm. Passing these control actions gives static pairs $\{\boldsymbol{s}^i_{\mathrm{static}}, \boldsymbol{a}^i_{\mathrm{static}}\}_{i=1}^{N}$ and their accompanying decay trajectories $\{\boldsymbol{s}^i_{\mathrm{decay}}, \boldsymbol{a}^i_{\mathrm{decay}}\}_{i=1}^{mN}$.

\begin{itemize}
\item \textbf{Step 1. [Learning the critical manifold].} We learn $\bar{\boldsymbol{s}}(\boldsymbol{a})$ as a nonlinear static map by regressing the static configurations onto their control actions,
\begin{equation}
\label{eq:learn_cm_RL}
\bar{\boldsymbol{s}}^{\star} = \operatorname*{arg\,min}_{\bar{\boldsymbol{s}}}\ \sum_{i=1}^{N} \big\| \boldsymbol{s}^i_{\mathrm{static}} - \bar{\boldsymbol{s}}(\boldsymbol{a}^i_{\mathrm{static}}) \big\|^2.
\end{equation}
We seek the critical manifold as a polynomial expansion of order $k_a$,
\begin{equation}
\label{eq:cm_poly_RL}
\bar{\boldsymbol{s}}(\boldsymbol{a}) = \sum_{0 \leq |\boldsymbol{\alpha}| \leq k_a} \mathbf{c}_{\boldsymbol{\alpha}}\, \boldsymbol{a}^{\boldsymbol{\alpha}},
\end{equation}
where $\boldsymbol{\alpha}$ is a multi-index, and $\mathbf{c}_{\boldsymbol{\alpha}}$ are the polynomial coefficients. The expansion is linear in $\mathbf{c}_{\boldsymbol{\alpha}}$, so the regression reduces to a linear least-squares problem with a closed-form solution. We can also use radial-basis and neural-network approximations for fitting $\bar{\boldsymbol{s}}(\boldsymbol{a})$ (see \citet{kaundinya2026precision} for details). 

\item \textbf{Step 2. [Learning the aSSM parametrization and reduced dynamics].} We center each decay about its static configuration to obtain $\{\boldsymbol{s}^i_{\mathrm{centered}}, \boldsymbol{a}^i_{\mathrm{decay}}\}_{i=1}^{mN}$. We later delay embed this data $p$ times through the map $\boldsymbol{\mu}$. The delays are chosen such that the condition $(p+1)n_s \geq 2d+1$ is satisfied. We collect the $m$ decays that share a static configuration, and perform an economy singular value decomposition to obtain chart maps $\{\mathbf{V}^i\}_{i=1}^{N}$, which we interpolate over $\boldsymbol{a}_{\mathrm{static}}$ or, when the local charts nearly coincide, use a single global chart. The data in the reduced coordinates is then obtained by $\boldsymbol{\mathcal{R}}^i = (\mathbf{V}^i)^{\mathrm T} \big[\boldsymbol{\mu}(\boldsymbol{s}^{i_1}_{\mathrm{centered}}), \dots, \boldsymbol{\mu}(\boldsymbol{s}^{i_m}_{\mathrm{centered}})\big]$ over the $m$ decays sharing the $i^{\text{th}}$ static configuration, and finite differencing this results in the velocity data $\dot{\boldsymbol{\mathcal{R}}}^i$.

We seek the leading-order aSSM parametrization $\mathbf{W}(\cdot)$ and reduced dynamics $\mathbf{R}(\cdot)$ as a polynomial expansion of the form,
\begin{align}
\label{eq:W_poly_RL}
\mathbf{W}(\mathbf{r}, \boldsymbol{s}) &= \sum_{1 \leq |\boldsymbol{\alpha}| \leq k^W_r} \sum_{0 \leq |\boldsymbol{\beta}| \leq k^W_a} \mathbf{W}_{\boldsymbol{\alpha}\boldsymbol{\beta}}\, \mathbf{r}^{\boldsymbol{\alpha}}\,\boldsymbol{s}^{\boldsymbol{\beta}}, \\
\label{eq:R_poly_RL}
\mathbf{R}(\mathbf{r}, \boldsymbol{s}) &= \sum_{1 \leq |\boldsymbol{\alpha}| \leq k^R_r} \sum_{0 \leq |\boldsymbol{\beta}| \leq k^R_a} \mathbf{R}_{\boldsymbol{\alpha}\boldsymbol{\beta}}\, \mathbf{r}^{\boldsymbol{\alpha}}\, \boldsymbol{s}^{\boldsymbol{\beta}},
\end{align}
where $\boldsymbol{\alpha}$ and $\boldsymbol{\beta}$ are multi-indices over $\mathbf{r}$ and $\boldsymbol{s}$ respectively, $k^W_r, k^W_a$ and $k^R_r, k^R_a$ denote the corresponding polynomial orders, and $\mathbf{W}_{\boldsymbol{\alpha}\boldsymbol{\beta}}$, $\mathbf{R}_{\boldsymbol{\alpha}\boldsymbol{\beta}}$ are the unknown coefficients. We fit these by minimizing the reconstruction and dynamics regression losses on the centered decay data, 
\begin{align}
\label{eq:learn_aSSM_RL}
\mathbf{W}^{\star} &= \operatorname*{arg\,min}_{\mathbf{W}}\ \sum_i \big\| \boldsymbol{S}^i_{\mathrm{centered}} - \mathbf{W}\big(\boldsymbol{\mathcal{R}}^i, \boldsymbol{s}^i_{\mathrm{static}})\big) \big\|^2, \\
\mathbf{R}^{\star} &= \operatorname*{arg\,min}_{\mathbf{R}}\ \sum_i \big\| \dot{\boldsymbol{\mathcal{R}}}^i - \mathbf{R}\big(\boldsymbol{\mathcal{R}}^i, \boldsymbol{s}^i_{\mathrm{static}})\big) \big\|^2, \nonumber
\end{align}
where $\boldsymbol{S}^i_{\mathrm{centered}} = \big[\boldsymbol{s}^{i_1}_{\mathrm{centered}}, \dots,\boldsymbol{s}^{i_m}_{\mathrm{centered}}\big]$. All  parameters concerning the aSSM-reduced model are chosen to minimize the normalized mean trajectory error on unseen test decay data.
\end{itemize}

\subsection{Offline model-based reinforcement learning}\label{sec:rl}

We learn a policy entirely from the learned dynamics model. The dynamics model is identified offline from the dataset $\mathcal{D}$ and is held fixed
thereafter. No interaction with the hardware takes place during policy optimization.
The policy $\boldsymbol{\pi}_{\theta}$ acts on the observation
\begin{equation}
    \mathbf{o}_k = \left\{ \boldsymbol{s}_k,\, {\boldsymbol{s}}_{k-1},\, \boldsymbol{a}_{k-1},\, \boldsymbol{\Gamma}_{k+1} \right\},
    \label{eq:observation}
\end{equation}
where $\boldsymbol{s}_k$ is the state obtained from passing control action $\boldsymbol{a}_{k-1}$ to the dynamics model and
$\boldsymbol{\Gamma}_{k+1}$ the target state to be reached. 
 The control action needed to reach the target state $\boldsymbol{\Gamma}_{k+1}$ is sampled as,
\begin{equation}
    \boldsymbol{a}_k \sim \boldsymbol{\pi}_{\theta}\!\left( \cdot \,\middle|\, \mathbf{o}_k \right).
    \label{eq:policy}
\end{equation}

We note, that for obtaining $\boldsymbol{o}_k$ requires $\boldsymbol{s}_{k}$ that needs to evaluated using the aSSM model in eq.~(\ref{eq:aSSM_discrete_RL}) or the autoregressive model eq.~(\ref{eq:ar_model}). The aSSM model requires the control action at $k-1$, whereas the autoregressive model requires past control actions based on the model's memory window. The aSSM model needs past states only to set an initial condition, but the dynamical rollouts are history-independent. The autoregressive model also needs a history of states to perform the rollout.

\begin{algorithm}
\caption{RL policy optimization by imagination on the learned dynamics model}
\label{algo:mbpo}
\begin{algorithmic}[1]
\Require Dynamics model either an aSSM model eq.~(\ref{eq:aSSM_discrete_RL}) or a neural-network-based model eq.~(\ref{eq:ar_model}) identified offline from $\mathcal{D}$ and held fixed
\Require Dataset $\mathcal{D}$, target distribution $\mathcal{T}$, horizon $T$, number of parallel imagined rollouts $N$
\State Initialize policy $\boldsymbol{\pi}_\theta$
\For{learning iterations $= 1, 2, \dots$}
    \State \textbf{Imagination:} sample $N$ initial conditions and target tracks, then roll out in parallel
    \For{$i = 1, \dots, N$ \textbf{in parallel}}
        \State Sample warm-start history and initial actions from $\mathcal{D}$, giving the initial condition of rollout $i$
        \State Sample a target track $\{\boldsymbol{\Gamma}^{(i)}_k\}_{k=1}^{T}$ from $\mathcal{T}$
    \EndFor
    \For{$k = 0, 1, \dots, T-1$}
        \For{$i = 1, \dots, N$ \textbf{in parallel}}
            \State Assemble $\mathbf{o}^{(i)}_k$ by Eq.~\eqref{eq:observation}
            \State Sample $\boldsymbol{a}^{(i)}_k \sim \boldsymbol{\pi}_\theta(\cdot|\mathbf{o}^{(i)}_k)$
            \State Advance one step to $\boldsymbol{s}^{(i)}_{k+1}$ using the dynamics model
            \State Evaluate reward $\rho^{(i)}_k$ by Eq.~\eqref{eq:reward}
        \EndFor
    \EndFor
    \State Update $\boldsymbol{\pi}_\theta$ on the batch of $N$ imagined rollouts using PPO
\EndFor
\end{algorithmic}
\end{algorithm}
\FloatBarrier

The target horizon length is $T$ steps. Thus, the model must perform $T$ rollouts to produce a prediction based on sampled control-action sequences. At each step of the rollout, we evaluate the reward, defined by
\begin{equation}
    \rho_k = \rho\big(\boldsymbol{s}_{k+1},\, \boldsymbol{a}_k,\, \boldsymbol{\Gamma}_{k+1}\big).
    \label{eq:reward}
\end{equation}
We provide Algorithm \ref{algo:mbpo} for training a policy that maximizes the above reward along the target track across different initial conditions.

\subsection*{Acknowledgements}

This work has been partially supported by grant $200021\_214908$ from the Swiss National Science
Foundation (SNF).

%\subsection*{Author Contributions}

\subsection*{Competing Interests}

All authors declare no financial or non-financial competing interests.

%% This article has accompanying supplementary information.

\clearpage
\section*{Extended Data}

\begin{edfigure}[H]
    \centering
    % Row: a (commanded vs. measured pressure), b (resulting response), side by side
    \begin{subfigure}[t]{0.49\linewidth}
        \centering
        \strut\par
        \includegraphics[width=\linewidth]{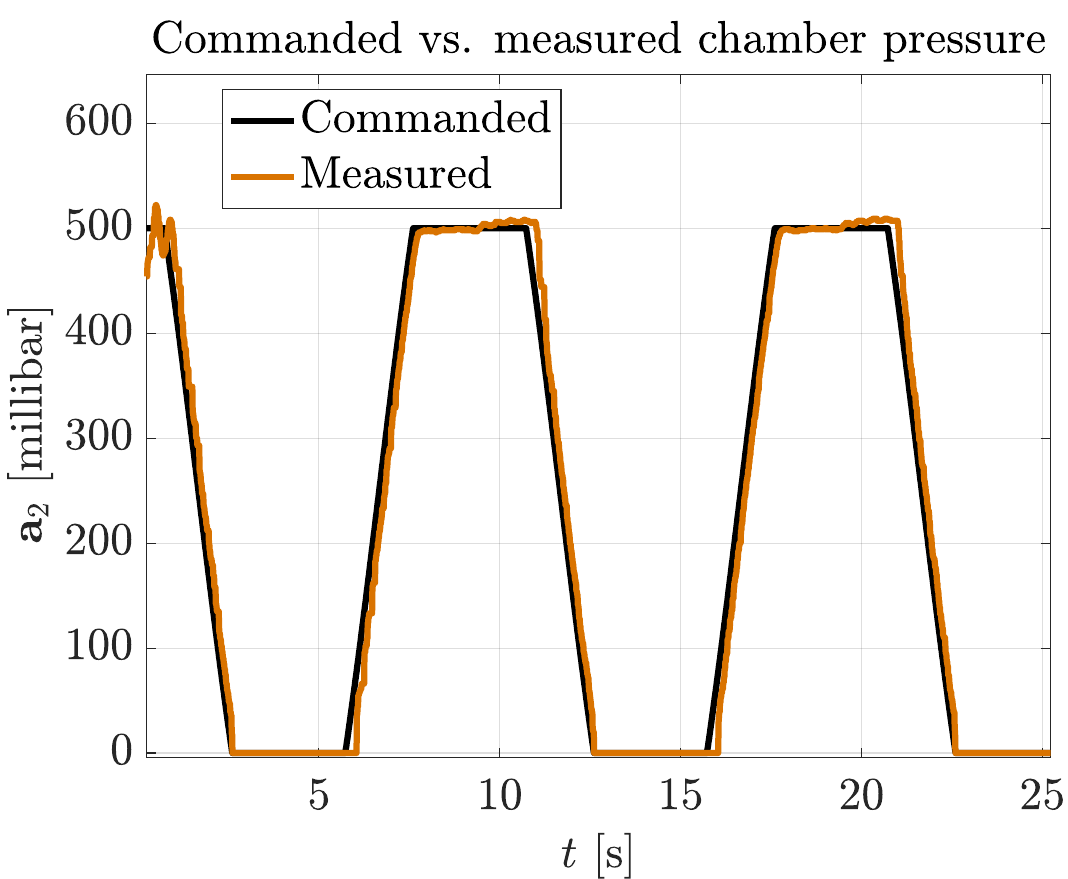}
        \caption{}
        \label{edfig:commanded}
    \end{subfigure}\hfill
    \begin{subfigure}[t]{0.49\linewidth}
        \centering
        \strut\par
        \includegraphics[width=\linewidth]{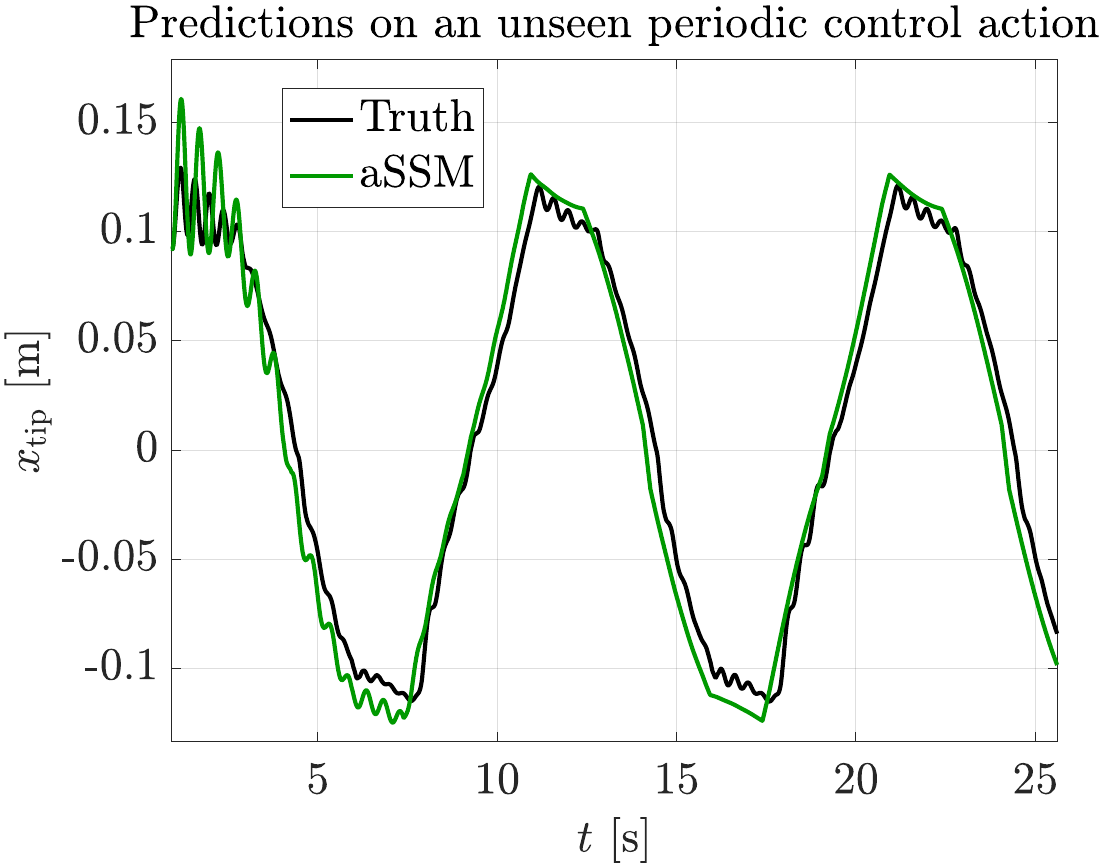}
        \caption{}
        \label{edfig:periodic_response}
    \end{subfigure}
    \caption{\textbf{Mismatch between commanded and measured pressure in SoPrA.}
    \textbf{a}, Time series of the control action $\boldsymbol{a}_2$ for the periodic
    case, with the pressure commanded by the user (black) and the pressure measured in
    the chamber (orange). \textbf{b}, The mismatch in \textbf{a} makes the measured
    response of the SoPrA arm (black) fluctuate about the periodic response that the
    aSSM model predicts (green).}
    \label{edfig:sopra_pressure}
\end{edfigure}

\begin{edfigure}[H]
    \centering
    \includegraphics[width=0.6\linewidth]{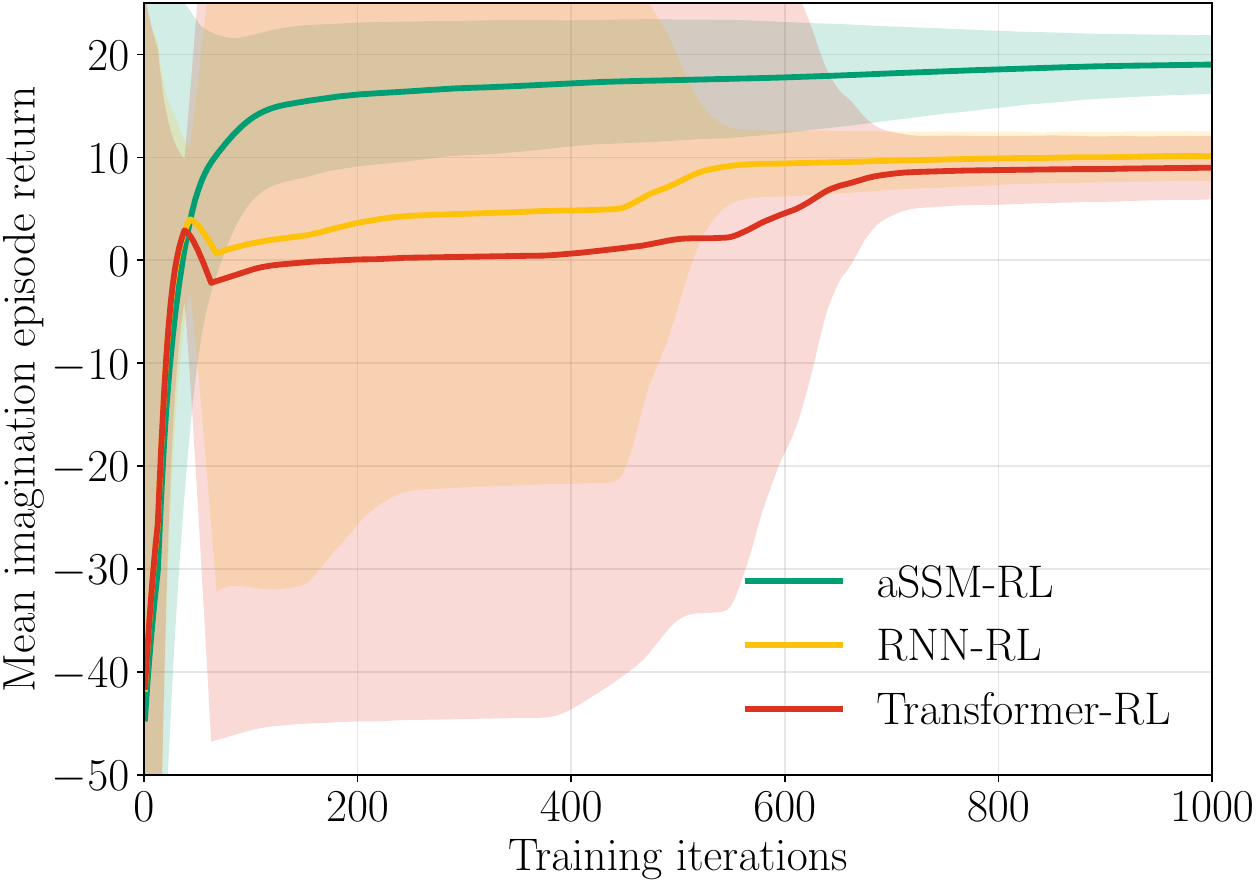}
    \caption{\textbf{[Convergence of model-based RL policies trained offline for closed-loop trajectory tracking]}. Plots for the evolution of the mean imagination episode return for aSSM-RL (green), RNN-RL (yellow), and Transformer-RL (red) as the policies are trained. The solid lines for each policy depict the average mean imagination episode return across hyperparameter sweeps, and the shaded region shows the deviation.}
    \label{edfig:rl_reward}
\end{edfigure}

\begin{edfigure}[tbp]
    \centering
    \begin{subfigure}[t]{\linewidth}
        \centering
        \makebox[\linewidth][c]{%
            \includegraphics[width=\linewidth]{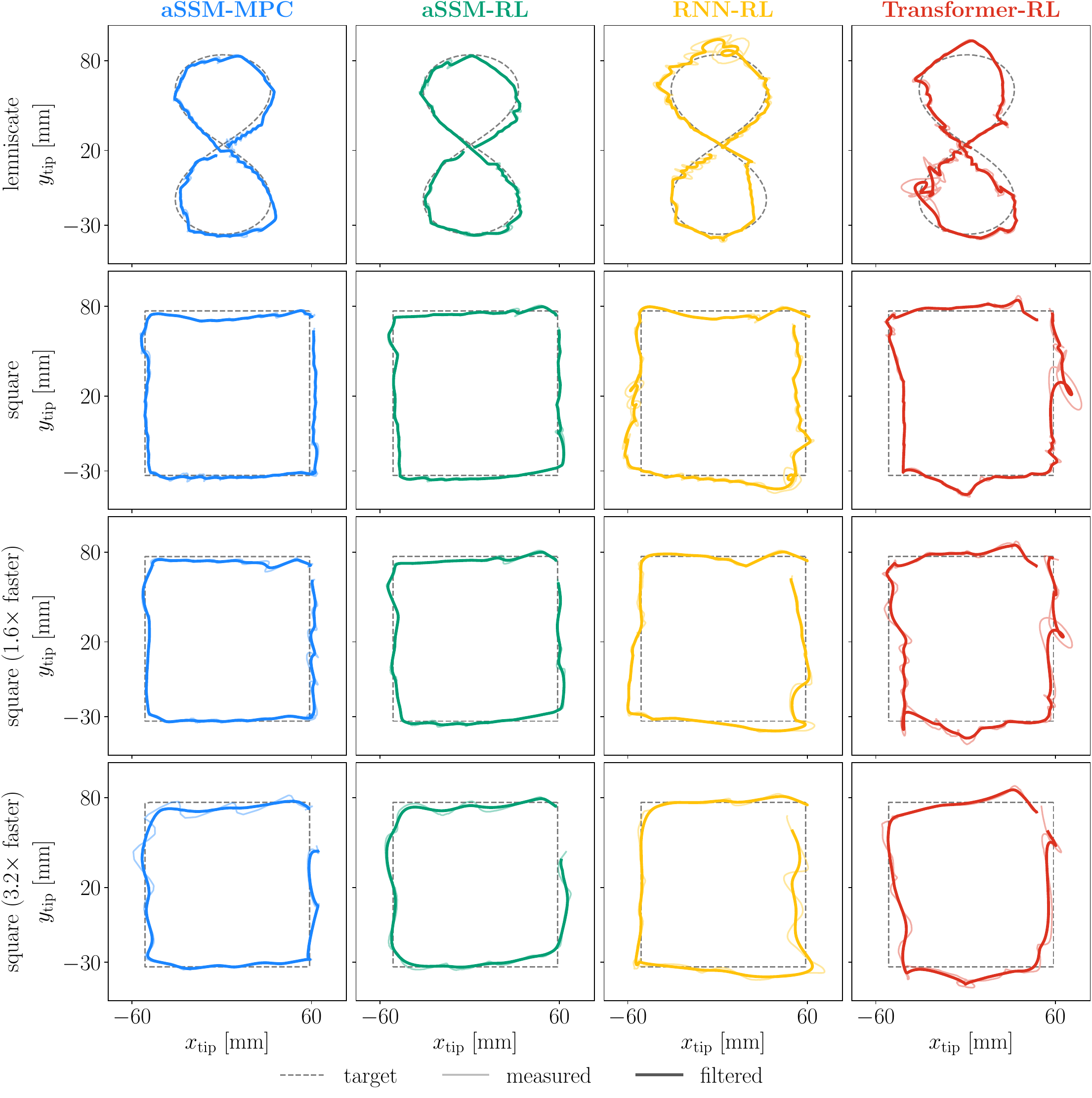}}
        \caption{}
        \label{fig:cloop_filter}
    \end{subfigure}

    \vspace{6pt}

    \begin{subfigure}[t]{\linewidth}
        \centering
        \makebox[\linewidth][c]{%
            \includegraphics[width=\linewidth]{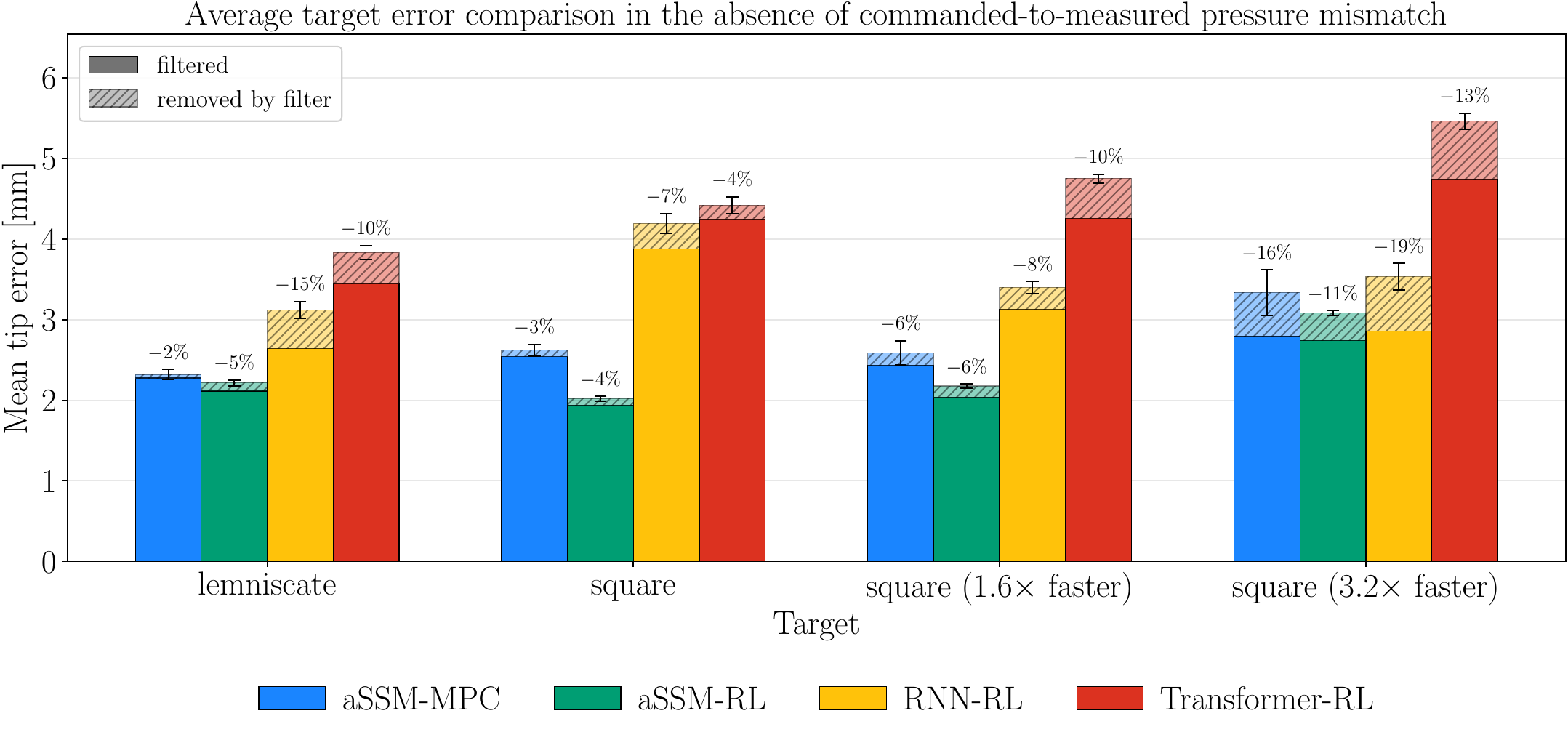}}
        \caption{}
        \label{fig:clooperror_filter}
    \end{subfigure}
    \caption{\textbf{[Closed-loop results with potential hardware upgrades]}. (a) Closed-loop results for four target tracks for aSSM-MPC (blue), aSSM-RL (green), RNN-RL (yellow), and Transformer-RL (red), with numerical filtering applied to the results to demonstrate the improvement expected when we use upgraded pressure valves. (b) Revised average tracking error compared across all the target tracks when commanded-to-measure pressure mismatch is accounted for numerically. The percentage number on top of each bar shows the extent of error removed by the numerical filter.}
    \label{edfig:cloop_upgrades}
\end{edfigure}

\end{document}